\documentclass[11pt]{article}

\newif\ifupdatepdf
\updatepdftrue

\newif\ifreview
\reviewfalse

\newif\ifpreprint
\preprintfalse

\ifreview
\ifpreprint
\PackageError{MyDocument}{ifreview and ifpreprint are true. This is an invalid combination}{Use \reviewtrue and \preprintfalse.}
\fi
\fi

\ifreview
\usepackage[review]{acl}
\else
\ifpreprint
\usepackage[preprint]{acl}
\else
\usepackage[final]{acl}
\fi
\fi

\usepackage{times}
\usepackage{latexsym}

\usepackage[T1]{fontenc}

\usepackage[utf8]{inputenc}

\usepackage{microtype}

\usepackage{inconsolata}

\usepackage{graphicx}

\usepackage[dvipsnames,table]{xcolor}
\usepackage{subcaption}
\usepackage{wrapfig}
\usepackage{booktabs}
\usepackage{amsfonts}
\usepackage{multirow}
\usepackage{longtable}
\usepackage{pifont}
\usepackage{lipsum}
\usepackage{mdframed}
\usepackage{soul}  %
\usepackage{tablefootnote}
\usepackage{listings}
\usepackage[most]{tcolorbox}
\usepackage{setspace}
\usepackage{amsmath}

\usepackage{tikz}
\usetikzlibrary{patterns}

\usepackage{minted}
\definecolor{minted@bgcolor}{rgb}{0.9, 0.8, 1.0}

\usepackage{array}
\newcolumntype{M}[1]{>{\centering\arraybackslash}m{#1}}
\newcolumntype{L}[1]{>{\raggedright\arraybackslash}m{#1}}
\newcolumntype{R}[1]{>{\raggedleft\arraybackslash}m{#1}}

\usepackage{enumitem}
\setlist[itemize]{leftmargin=*,itemsep=2pt} %

\newcommand{\symfootnotetext}[2]{%
  \setcounter{footnote}{#1}%
  \renewcommand{\thefootnote}{\fnsymbol{footnote}}%
  \footnotetext{#2}%
}

\newtcolorbox{takeawaybox}[1]{
    colback=gray!10,
    colframe=blue!20,
    arc=4pt,
    left=2pt,
    right=2pt,
    top=2pt,
    bottom=2pt,
    boxsep=2pt,
    title={#1},
    coltitle=black,
    fonttitle=\bfseries,
    center title,
    toptitle=1pt
}

\renewcommand{\paragraph}[1]{\vskip .4em \noindent {\bf #1}}

\title{Evaluating LLM-Simulated Conversations in Modeling Inconsistent and Uncollaborative Behaviors in Human Social Interaction}

\newcommand{\intern}{\textsuperscript{\dag}}
\newcommand{\msft}{\textsuperscript{1}}
\newcommand{\psu}{\textsuperscript{2}}
\newcommand{\usc}{\textsuperscript{3}}

\author{Ryo Kamoi\psu\intern, Ameya Godbole\usc\intern, Binglin Zhou\psu, Xiaoxin Lu\psu,\\
{\bf Longqi Yang\msft, Rui Zhang\psu, Mengting Wan\msft, Pei Zhou\msft } \\
\msft Microsoft Corporation, \psu Penn State University, \usc University of Southern California \\
\texttt{\{ryokamoi, rmz5227\}@psu.edu, \{mengtwan, zhoupei\}@microsoft.com}
}

\begin{document}

\let\oldaddcontentsline\addcontentsline
\renewcommand{\addcontentsline}[3]{}

\maketitle

\ifreview
\else
\symfootnotetext{2}{Work done during an internship at Microsoft Corporation.}
\fi

\newcommand{\cocoeval}{\textsc{CoCoEval}}
\newcommand{\cocoevalbench}{\textsc{CoCoEvalBench}}

\begin{abstract}

Simulating human conversations using large language models (LLMs) has emerged as a scalable methodology for modeling human social interaction. This paper reconsiders the evaluation of simulated conversations by explicitly recognizing that human conversations inherently involve inconsistent and uncollaborative behaviors, such as misunderstandings and interruptions. Since these behaviors contribute to the complexity of human social interaction, we argue that LLM-simulated conversations should reproduce them at frequencies comparable to those observed in human conversations. To support a detailed and interpretable evaluation of these behaviors, we introduce \cocoeval{}, a framework consisting of an evaluation scheme based on turn-level detection of 10 types of inconsistent and uncollaborative behaviors and a benchmark for simulating conversations in professional scenarios involving collaboration and conflict. Using \cocoeval{}, we compare human conversations with those simulated by GPT-4.1, GPT-5.1, and Claude~Opus~4. The results show that (1)~LLM-simulated conversations exhibit far fewer inconsistent and uncollaborative behaviors than human conversations under vanilla prompting, and (2)~prompt engineering and supervised fine-tuning do not provide reliable control over these behaviors, often leading to the overproduction of specific behaviors. \cocoeval{} identifies gaps between human and LLM-simulated conversations that are not captured by conventional evaluation based on conversation-level Likert scales, raising concerns about the use of LLMs as proxies for human social interaction.
\ifreview
The benchmark, model outputs, and prompts are in the supplementary material.
\else
\fi
\end{abstract}

\begin{figure*}[t!]
    \centering
    \includegraphics[width=\linewidth]{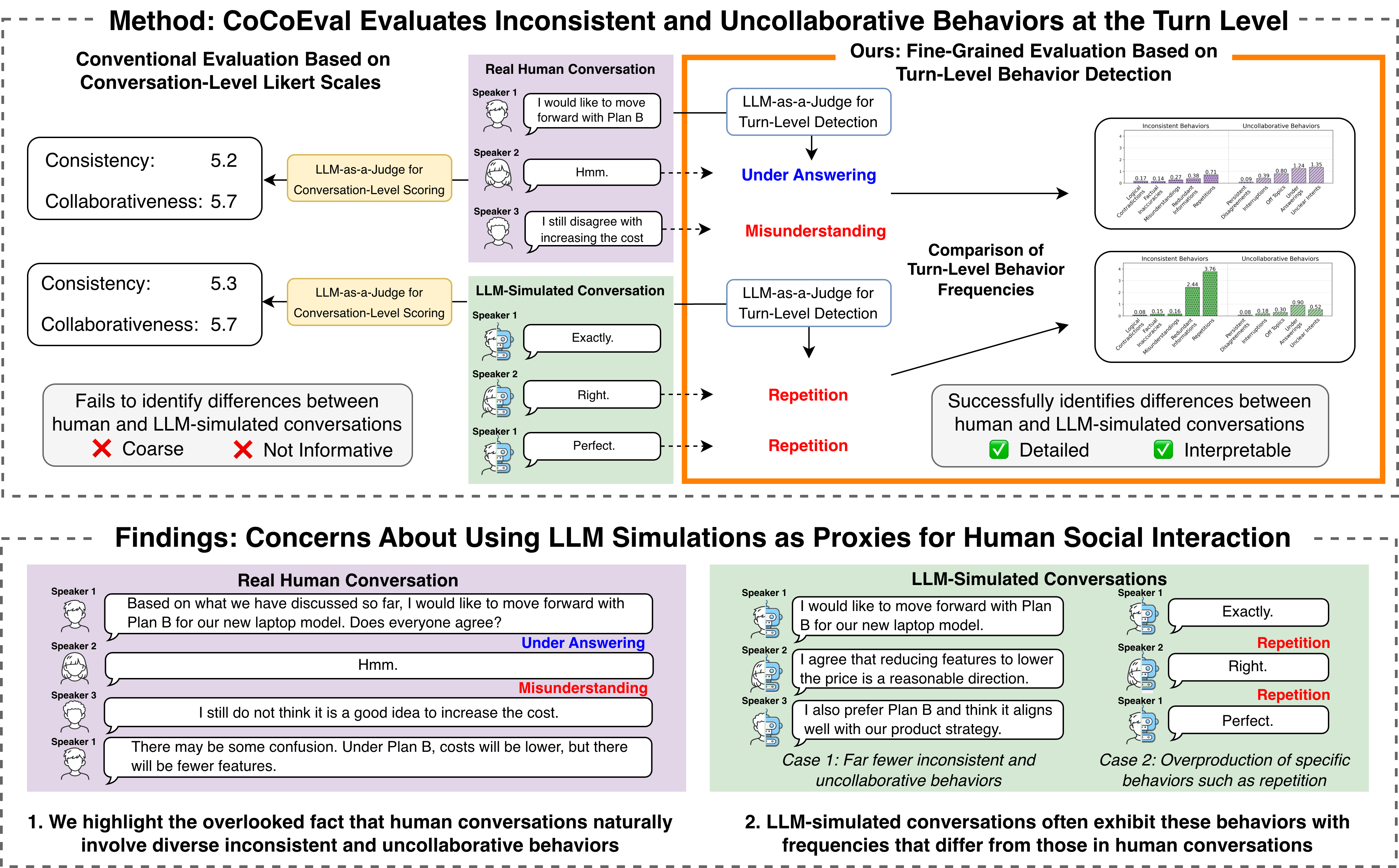}
    \caption{{\bf Method (top):}~This paper introduces \cocoeval{}, a framework for evaluating inconsistent and uncollaborative behaviors in LLM-simulated conversations. \cocoeval's fine-grained evaluation scheme detects inconsistent and uncollaborative behaviors at the turn level~(right). Compared to conventional evaluation based on conversation-level Likert scales~(left), our method provides more interpretable and detailed analysis, better capturing differences in conversational properties between humans and LLMs.
    {\bf Findings (bottom):}~Our analysis highlights the overlooked fact that human conversations inherently involve inconsistent and uncollaborative behaviors~(left). \cocoeval{} also reveals that LLM-simulated conversations fail to reproduce these behaviors in human conversations, often containing far fewer of them or overproducing specific behaviors, depending on the simulation setup~(right).
    }
    \label{fig:overview}
\end{figure*}

\section{Introduction}

Simulating human conversations by generating utterances for all participants using large language models~(LLMs) has recently emerged as a scalable methodology for modeling human social interaction, with potential applications in prototyping social platforms such as social media~\citep{park2023generative}, testing hypotheses in social science~\citep{manning2024automated}, and helping decision-makers forecast how people may respond to new policies~\citep{sreedhar2024simulating}. However, it remains an open question to what extent LLMs can reproduce human behaviors and serve as a reliable proxy for human social interaction, and how we should evaluate the proximity of LLM-simulated conversations to human ones.

This paper reconsiders the evaluation of LLM-simulated conversations by explicitly recognizing that human conversations are not perfectly consistent or collaborative and inherently involve inconsistent behaviors, such as misunderstandings and repetition, as well as uncollaborative behaviors, such as interruptions and disagreements~\citep{grice1975logic, schegloff1992repair, pickering2004toward}. While these properties of human conversation are well documented, they have been largely overlooked in the evaluation of LLM-simulated conversations. We argue that reproducing the frequencies of inconsistent and uncollaborative behaviors is necessary for modeling human social interaction using LLM-simulated conversations, since these behaviors contribute to the complexity of human social interactions by giving rise to downstream social behaviors such as recovering from misunderstandings or persuading other participants. This work addresses two research gaps. (1)~First, prior work often implicitly assumes that higher consistency and collaborativeness correspond to better simulation quality~\citep{zhou2024sotopia, wang2025multiparty}. This assumption is misaligned with the goal of simulating human social interaction, which is not perfectly consistent or collaborative. (2)~Second, prior work typically evaluates simulated conversations using conversation-level Likert scales~\citep{park2023generative, zhou2024sotopia}, such as consistency and collaborativeness rated on a 1--10~scale, which does not provide a detailed or interpretable analysis.

We introduce \cocoeval{}, an evaluation framework comprising an evaluation scheme and a benchmark designed to facilitate detailed and interpretable analysis of inconsistent and uncollaborative behaviors in human and LLM-simulated conversations. First, we propose a fine-grained evaluation scheme that detects inconsistent and uncollaborative behaviors at the turn level. We define 10~types of behaviors based on taxonomies for dialogue generation~\citep{higashinaka-etal-2021-integrated, wang-etal-2023-chatgpt-defend}, as in Table~\ref{tab:behavior-definition}. A successful simulation should include these behaviors at frequencies comparable to those in human conversations. Second, we introduce \cocoevalbench, a benchmark to evaluate LLMs on simulating 30-turn conversation continuations given metadata and history. It is created from real-world human conversations in collaborative or conflicting situations in QMSum~\citep{zhong-etal-2021-qmsum}, NCPC~\citep{ncpc2024}, SIM~\citep{fan2024topicconversation}, and IQ2~\citep{zhang-etal-2016-conversational}, spanning business, academic, and governmental meetings as well as debates.

\begin{table*}[t]
\centering
\scriptsize
\begin{tabular}{M{.1\linewidth}M{.095\linewidth}M{.33\linewidth}M{.37\linewidth}}
\toprule
 & Name & Definition & Example \\
\midrule
\multirow{12}{\linewidth}{\centering Inconsistent Behaviors}
& Logical Contradiction & Utterances that contradict earlier statements by the same speaker or the speaker's metadata. & One participant says ``The remote control runs on a single AA battery,'' and later says ``It requires two AA batteries.'' \\
\cmidrule{2-4}
& Factual Inaccuracy & Utterances that contain factually incorrect statements. & A participant claims that infrared signals can pass through walls. \\
\cmidrule{2-4}
& Misunder-standing & Utterances that reflect a misinterpretation of information provided by other participants. & A participant states that the remote control uses only a touch screen, but another later assumes that it has physical buttons. \\
\cmidrule{2-4}
& Redundant Information & Utterances that contain unnecessarily lengthy content and add little new informational value. & A participant gives a long explanation about previous projects without adding new information. \\
\cmidrule{2-4}
& Repetition & Utterances that unnecessarily repeat information stated earlier in the conversation. & One participant says ``I will bring the documents,'' and another asks ``So you will bring the documents, right?'' \\
\midrule
\multirow{12}{\linewidth}{\centering Uncollaborative Behaviors}
& Persistent Disagreement & Utterances that repeatedly reject others' positions despite established consensus or supporting evidence. & A participant continues to reject a proposed plan even after others agree and provide supporting evidence for it. \\
\cmidrule{2-4}
& Interruption & Utterances that disrupt the conversation by cutting off another participant. & A participant interrupts another speaker mid-explanation and leaves the utterance unfinished, like ``... the key point is''. \\
\cmidrule{2-4}
& Off Topic & Utterances that deviate substantially from the current conversational topic. & During the middle of a project meeting discussion, a participant shares unrelated personal stories. \\
\cmidrule{2-4}
& Under Answering & Utterances that give evasive responses and fail to address preceding questions. & When asked ``Will this policy increase taxes?'', a participant replies ``We are focused on supporting working families.'' \\
\cmidrule{2-4}
& Unclear Intent & Utterances that are ambiguous or fail to clearly convey the speaker's intentions. & When asked whether to agree or disagree, a participant responds only with ``Maybe.'' \\
\bottomrule
\end{tabular}
\caption{Definitions of inconsistent and uncollaborative behaviors used in the fine-grained evaluation in \cocoeval{}.}
\label{tab:behavior-definition}
\end{table*}

Using \cocoeval, we evaluate GPT-4.1, GPT-5.1, and Claude~Opus~4 in widely used conversation simulation setups. \cocoeval{} reveals discrepancies between human and LLM-simulated conversations in inconsistent and uncollaborative behaviors, as in Figure~\ref{fig:overview}. First, under vanilla prompting, LLMs generate conversations with far fewer inconsistent and uncollaborative behaviors than human conversations~(\S\ref{sec:results-vanilla-prompting}). Second, prompt engineering does not provide reliable control over these behaviors. LLMs overproduce these behaviors when explicitly instructed to include them at frequencies observed in human conversations~(\S\ref{sec:results-prompt-engineering}). Third, supervised fine-tuning on human conversations results in the overproduction of a narrow range of behaviors such as repetition and redundancy~(\S\ref{sec:results-finetuning}).

In summary, our analysis of inconsistent and uncollaborative behaviors in LLM-simulated conversations using \cocoeval{} shows that the properties of simulated conversations are highly sensitive to simulation settings, and that none of the evaluated settings reproduces the frequencies of these behaviors observed in human conversations. The results also demonstrate an advantage of \cocoeval{}'s evaluation scheme based on turn-level behavior detection, which identifies differences between human and LLM-simulated conversations that conventional conversation-level Likert scale evaluation fails to capture~(\S\ref{sec:discussion}), as illustrated in Figure~\ref{fig:overview}. Below, we summarize our contributions:
\begin{itemize}
    \item We are the first to analyze LLM-simulated conversations with a focus on inconsistent and uncollaborative behaviors, which are inherent in human conversations and should be reproduced to simulate human social interaction.
    \item We introduce \cocoeval{}, a framework for analyzing inconsistent and uncollaborative behaviors in human and simulated conversations. It consists of an evaluation scheme designed to provide detailed and interpretable analysis, along with a benchmark featuring professional scenarios.
    \item Our analysis using \cocoeval{} reveals that LLM-simulated conversations often exhibit inconsistent and uncollaborative behaviors at frequencies that differ from those in human conversations, raising concerns about their use as proxies for human social interaction.
\end{itemize}

\section{Related Work}

We study the simulation of multi-party conversations in which LLMs simulate all participants~\citep{park2023generative, manning2024automated, sreedhar2024simulating, zhou2024sotopia}. This setting has been studied for modeling human social interaction beyond simulating a single individual, in contrast to situations where LLMs simulate a single participant~\citep{elhilali2025large} or human-LLM conversations~\citep{zheng-etal-2023-augesc, wu2025collabllm}.

Consistency and collaborativeness in this paper cover most of the commonly used evaluation dimensions in conversation simulation, including coherence with conversation history~\citep{maharana-etal-2024-evaluating, chu2024cohesive}, alignment with persona and role~\citep{frisch-giulianelli-2024-llm, abdulhai2025consistently}, adherence to social norms~\citep{li-etal-2023-normdial}, and social interaction capabilities~\citep{zhou2024sotopia, piedrahita2025corrupted}, while a few dimensions, such as behavioral diversity~\citep{pang-etal-2020-towards}, are not covered. Evaluation in prior work commonly relies on conversation-level Likert scales, whereas we introduce an evaluation scheme based on turn-level behavior detection to enable more detailed and interpretable evaluation.

\section{\cocoeval}

We introduce \cocoeval{}, an evaluation framework for analyzing LLM-simulated conversations with a focus on inconsistent and uncollaborative behaviors, which are inherent to human conversations~\citep{schegloff1992repair, grice1975logic, pickering2004toward} and contribute to the complexity of human social interaction. \cocoeval{} comprises an evaluation scheme~(\S\ref{sec:evaluation-scheme}) and a benchmark~(\S\ref{sec:cocoevalbench}) designed to facilitate detailed and interpretable comparison between human conversations and LLM-simulated conversations with respect to these behaviors. We implement the evaluation using LLM-as-a-Judge to support scalable evaluation~(\S\ref{sec:llm-as-a-judge-agreement}).

The definitions of consistency and collaborativeness in \cocoeval{} follow prior discussions in conversation evaluation. {\bf Consistency} includes coherence with previous utterances, adherence to the specified persona, and correctness with respect to world knowledge~\citep{park2023generative, frisch-giulianelli-2024-llm, abdulhai2025consistently}, as well as the avoidance of unnecessary redundancy and repetition~\citep{higashinaka-etal-2021-integrated, liu2025promediate}. {\bf Collaborativeness} is the extent to which a speaker's contributions support shared objectives and help other participants achieve their goals~\citep{zhou2024sotopia, piatti2024cooperate, piedrahita2025corrupted, nonomura2025who}, without delaying or impeding progress and agreement.

\subsection{Evaluation Scheme} \label{sec:evaluation-scheme}

\cocoeval{} adopts two evaluation schemes. The first is the overall evaluation based on Likert scale scores, which is a common evaluation approach in conversation simulation~\citep{zhou2024sotopia, wang2025multiparty}. However, it is coarse-grained and unsuitable for detailed analysis. To provide a more detailed and interpretable evaluation, we introduce the fine-grained evaluation that detects inconsistent and uncollaborative behaviors at the turn level.

\paragraph{Overall evaluation} assigns two conversation-level Likert scale scores, one for consistency and one for collaborativeness, each on a 1--10 scale. We provide definitions and scoring rubrics describing the criteria for each score range~(Table~\ref{prompt:overall-evaluation}).

\paragraph{Fine-grained evaluation} detects 10 types of inconsistent and uncollaborative behaviors at the turn level. We derive 10 types of inconsistent and uncollaborative behaviors shown in Table~\ref{tab:behavior-definition} by adopting nine types of behaviors from evaluation taxonomies in dialogue generation~\citep{higashinaka-etal-2021-integrated, mahajan-etal-2022-towards}, and we add ``persistent disagreement'' to capture uncollaborative behavior in negotiation and debates~\citep{wang-etal-2023-chatgpt-defend, liu2025promediate}. For each category, we detect turns that exhibit the corresponding behaviors and count the number of occurrences in each conversation.

\begin{figure}[t]
    \centering
    \includegraphics[width=\linewidth]{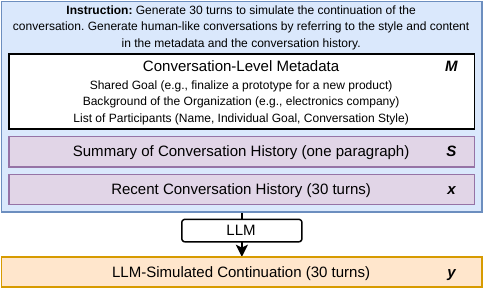}
    \caption{Task in \cocoevalbench.}
    \label{fig:continuation-task}
\end{figure}

\subsection{\cocoevalbench} \label{sec:cocoevalbench}

We introduce \cocoevalbench, a benchmark for evaluating LLMs on the conversation continuation task, in which models simulate continuations given conversational metadata and history~(Figure~\ref{fig:continuation-task}). It covers real-world collaborative or conflicting scenarios spanning business, academic, and governmental meetings, as well as debates. Compared to prior work~\citep{zhou2024sotopia, bianchi2024how}, we focus on longer conversations with complex social interactions in professional scenarios. Accordingly, we design the task to require simulating 30-turn continuations, which are longer than the conversations used in most prior work.

\paragraph{Task.}
The input consists of metadata~$M$ containing the conversation goals and participant information, the 30~preceding turns~$x = [x_{-30}, \ldots, x_{-1}]$, and a paragraph-length summary~$S$ of the conversation prior to~$x$. Each turn in~$x$ is represented as a speaker-utterance pair~$x_i = (p_i, u_i)$. Given this input, LLMs are instructed to generate 30 continuation turns~$y = [y_1, \ldots, y_{30}] \sim \mathrm{LLM}(M, S, x)$, where each turn is a pair of a speaker and an utterance. Our benchmark includes a human continuation as a reference~$r = [r_1, \ldots, r_{30}]$. We adopt this setup to support direct comparison between LLMs and humans. Since the simulated continuation~$y$ and the human reference continuation~$r$ share the same context~$(M, S, x)$, they are expected to reflect similar frequencies of inconsistent and uncollaborative behaviors that are coherent with the metadata and history, and substantial deviations in these behaviors indicate simulation failures.

\paragraph{Benchmark creation.}
We create \cocoevalbench{} from real-world human conversations in which multiple participants pursue shared or conflicting goals in professional scenarios involving complex human social interaction. Specifically, we use six public conversation datasets:~business meetings in QMSum~Product, academic meetings in QMSum~Academic, governmental meetings in QMSum Committee~\citep{zhong-etal-2021-qmsum} and NCPC~\citep{ncpc2024}, and debates in SIM~\citep{fan2024topicconversation} and IQ2~\citep{zhang-etal-2016-conversational}.
To create the conversation continuation task, we first use o4-mini to extract information from conversations in the source datasets to generate metadata~$M$. Next, we randomly select a start point for the conversation continuation task and extract the preceding 30 turns as history~$x$ and the following 30 turns as reference continuation~$r$. We then generate a paragraph-length summary $S$ of the conversation preceding the 30-turn history~$x$ using o4-mini.

\paragraph{Statistics.}
Table~\ref{tab:statistics} shows the statistics of \cocoevalbench. The evaluation set includes 300~conversations from six datasets, and the training set consists of 1,450~conversations from four datasets.

\begin{table}[t]
\setlength{\tabcolsep}{2.3pt}
\scriptsize
\centering
\begin{tabular}{L{.18\linewidth}lM{.1\linewidth}rr}
\toprule
\multicolumn{1}{c}{Category} & \multicolumn{1}{c}{Source Dataset} & \# Data Points & \multicolumn{1}{M{.17\linewidth}}{Average \#~Participants} &
\multicolumn{1}{M{.18\linewidth}}{Average \#~Tokens/Turn} \\
\midrule
\multicolumn{5}{c}{{\bf Evaluation Set}} \\
\midrule
Business
& QMSum Product & 50 & 4.0 & 8.0 \\
\midrule
Academic
& QMSum Academic & 50 & 5.7 & 10.0 \\
\midrule
\multirow{2}{\linewidth}{Governmental}
& QMSum Committee & 50 & 18.5 & 66.3 \\
& NCPC & 50 & 25.8 & 50.4 \\
\midrule
\multirow{2}{\linewidth}{Debates}
& SIM & 50 & 4.0 & 16.2 \\
& IQ2 & 50 & 10.1 & 41.3 \\
\midrule
\midrule
\multicolumn{5}{c}{{\bf Training Set}} \\
\midrule
Business
& QMSum Product & 500 & 4.0 & 9.3 \\
\midrule
Governmental
& QMSum Committee & 300 & 18.6 & 73.3 \\
\midrule
\multirow{2}{\linewidth}{Debates}
& SIM & 300 & 4.0 & 16.4 \\
& IQ2 & 350 & 10.7 & 47.3 \\
\bottomrule
\end{tabular}
\caption{Statistics of \cocoevalbench.}
\label{tab:statistics}
\end{table}

\subsection{LLM-as-a-Judge} \label{sec:llm-as-a-judge-agreement}

To support scalable evaluation, we implement the evaluation schemes using LLM-as-a-Judge. To validate its use as an automated proxy for human evaluation, we collect evaluations from five annotators on human and LLM-simulated conversations for randomly selected cases from \cocoevalbench{} and assess their agreement with LLM-as-a-Judge. We use OpenAI o4-mini~\citep{o3-and-o4-mini} as LLM-as-a-Judge in Section~\ref{sec:results} because of its reasonable agreement with human evaluation. Appendix~\ref{appendix:llm-as-a-judge-agreement} provides detailed settings and results using GPT-5.1 Instant and High~\citep{gpt-5-1} and Claude Opus 4~\citep{claude4} as LLM-as-a-Judge.

\paragraph{Overall evaluation.}
The five annotators evaluate the consistency and collaborativeness of 54~conversations by humans and GPT-4.1 simulation on a 1--10~scale. The average leave-one-out Spearman's rank correlation coefficients between human evaluation and LLM-as-a-Judge~(o4-mini) are 0.546 for consistency and 0.586 for collaborativeness. The inter-human agreement, measured by the average leave-one-out Spearman's rank correlation, is 0.521 and 0.441, respectively. LLM-as-a-Judge shows agreement with human evaluation comparable to the inter-human agreement, supporting its use as an automated proxy.

\begin{table}[t]
\scriptsize
\centering
\setlength{\tabcolsep}{3pt}
\begin{tabular}{lrrrr}
\toprule
\multirow{2.7}{*}{Category} &
\multirow{2.7}{*}{\shortstack{Inter-human\\Spearman's $\rho$}} &
\multirow{2.8}{*}{\shortstack{Human vs.\\o4-mini\\Spearman's $\rho$}} &
\multicolumn{2}{c}{\# Detected Turns} \\
\cmidrule(lr){4-5}
 & & &
Human & o4-mini \\
\midrule
Logical Contradiction & 0.211\phantom{$^*$} & 0.044\phantom{$^*$} & 6.0 & 4 \\
Factual Inaccuracy & 0.256\phantom{$^*$} & 0.253\phantom{$^*$} & 2.4 & 2 \\
Misunderstanding & 0.001\phantom{$^*$} & 0.208$^*$ & 4.6 & 6 \\
Redundant Information & 0.200$^*$ & 0.185\phantom{$^*$} & 10.4 & 12 \\
Repetition & 0.370$^*$ & 0.613$^*$ & 18.4 & 23 \\
\midrule
Persistent Disagreement & 0.358\phantom{$^*$} & 0.315$^*$ & 4.4 & 2 \\
Interruption & 0.728$^*$ & 0.673$^*$ & 33.4 & 12 \\
Off Topic & 0.423$^*$ & 0.478$^*$ & 11.6 & 15 \\
Under Answering & 0.356$^*$ & 0.661$^*$ & 16.8 & 38 \\
Unclear Intent & 0.212$^*$ & 0.672$^*$ & 9.8 & 42 \\
\bottomrule
\end{tabular}
\caption{Agreement between human evaluation~(5~annotators) and LLM-as-a-Judge~(o4-mini), as well as inter-human agreement~(leave-one-out Spearman's $\rho$), in the number of detected behaviors for \cocoeval's fine-grained evaluation across 1,080~turns in 36~conversations. $^*$:~$p < 0.05$~(one-sided bootstrap).}
\label{tab:finegrained-annotation-number-comparison}
\end{table}

\paragraph{Fine-grained evaluation.}
The five annotators detect 10 types of inconsistent and uncollaborative behaviors at the turn level across 1,080~turns from 36~conversations by humans and GPT-4.1 simulation. Table~\ref{tab:finegrained-annotation-number-comparison} reports the average leave-one-out Spearman's rank correlations between human evaluation and LLM-as-a-Judge~(o4-mini), as well as inter-human agreement measured in the same way, for the number of detected behaviors in each conversation. The results show statistically significant inter-human and human-LLM~(o4-mini) agreement for most behavior types that occur more than 10~times in the evaluated 1,080~turns. Specifically, LLM-as-a-Judge~(o4-mini) shows statistically significant correlations with human annotations for detecting all types of uncollaborative behaviors, as well as Repetition and Misunderstanding, while it shows positive but relatively small correlations for the remaining three inconsistent behaviors. These inconsistent behaviors, except for Redundant Information, are rare in the evaluated domains and therefore do not substantially affect our analysis, although further analysis is needed when applying the framework to domains with more frequent inconsistent behaviors. The table also presents the total number of detected behaviors in 1,080~turns, showing that LLM-as-a-Judge~(o4-mini) largely captures the overall trends observed in human evaluation, although it tends to overdetect Under Answering and Unclear Intent.

\section{Conversation Simulation Setup}

We evaluate widely used conversation simulation settings:~off-the-shelf LLMs with prompt engineering and LLMs fine-tuned on human conversations. We also study how simulation parameters influence inconsistent and uncollaborative behaviors.

\paragraph{LLMs.}
We evaluate conversations simulated by GPT-4.1~\citep{gpt-4-1}, GPT-5.1~Instant~\citep{gpt-5-1}, and Claude~Opus~4~\citep{claude4}.

\renewenvironment{quote}
  {\list{}{\leftmargin=1em \rightmargin=1em \topsep=.5em \parsep=0pt \itemsep=0pt}\item\relax}
  {\endlist}

\paragraph{Prompt.}
We evaluate two types of instructions to assess the influence of prompt engineering. First, in {\bf vanilla prompting}~(Table~\ref{prompt:generation-system}), we instruct LLMs to generate {\it ``human-like conversations by referring to the conversation history''} and {\it ``mimic the communication style and content in the metadata and the conversation history.''} Second, we introduce {\bf taxonomy-guided prompting}~(Table~\ref{prompt:taxonomy-guided}) that explicitly instructs LLMs to incorporate inconsistent and uncollaborative behaviors at frequencies comparable to those observed in the conversation history, and provides the list of the 10 behavior types in Table~\ref{tab:behavior-definition} as follows:
\begin{quote}
\it\fontsize{10pt}{10pt}\selectfont
Human conversations may involve inconsistent and uncollaborative behaviors [...] The following list outlines representative inconsistent and uncollaborative behaviors: [...]
\end{quote}

\paragraph{Generating multiple turns per LLM call.}
Generating multiple consecutive turns in a single LLM call is computationally efficient, but generating one turn at a time better reflects the turn-by-turn nature of human conversations~\citep{zhou-etal-2024-real}. To examine the impact of these generation strategies, we evaluate settings that generate 1, 5, and 30 turns per call, as illustrated in Figure~\ref{fig:generating-multiple-turns}. For Claude~Opus~4, we only evaluate the settings generating 1 or 5 turns per call, as it frequently reaches its maximum output length when generating 30 turns per call.

\paragraph{SFT on human conversations.}
Supervised fine-tuning is a popular approach for aligning LLMs for conversation simulation~\citep{wang2025multiparty, chuang2025debate}. We fine-tune GPT-4.1 on human conversations from the training set of \cocoevalbench. We fine-tune three models that generate 1, 5, and 30 turns per call, respectively.

\section{Results} \label{sec:results}

We compare human and LLM-simulated conversations using \cocoeval{} on the conversation continuation task in \cocoevalbench, evaluated with LLM-as-a-Judge~(o4-mini). Our analysis addresses the following research questions.

\paragraph{RQ1:}
How much do {\bf human conversations} exhibit inconsistent and uncollaborative behaviors?~(\S\ref{sec:results-human-conversations})

\paragraph{RQ2:}
With {\bf vanilla prompting}, how well do off-the-shelf LLMs simulate inconsistent and uncollaborative behaviors in human conversations?~(\S\ref{sec:results-vanilla-prompting})

\paragraph{RQ3:}
Can {\bf prompt engineering} control inconsistent and uncollaborative behaviors in conversations simulated by off-the-shelf LLMs?~(\S\ref{sec:results-prompt-engineering})

\paragraph{RQ4:}
Can {\bf supervised fine-tuning} enable LLMs to reproduce inconsistent and uncollaborative behaviors in human conversations?~(\S\ref{sec:results-finetuning})

\paragraph{RQ5:}
Does {\bf generating multiple turns per LLM call} affect inconsistent and uncollaborative behaviors in LLM-simulated conversations?~(\S\ref{sec:results-multiple-turns})

\begin{table}[t]
\setlength{\tabcolsep}{3.5pt}
\scriptsize
\centering
\begin{tabular}{cM{.26\linewidth}M{.15\linewidth}M{.2\linewidth}M{.2\linewidth}}
\toprule
 &  & \# Turns Per~Call & Consistency & Collaborati-veness \\
\midrule
\multicolumn{2}{c}{Human}
   & -- & \cellcolor[HTML]{CBE8B4}5.2 {\scriptsize [4.8, 5.6]} & \cellcolor[HTML]{CBE8B4}5.7 {\scriptsize [5.3, 6.0]} \\
\midrule
\multirow{11}{*}{\rotatebox{90}{GPT-4.1}}
& \multirow{3}{\linewidth}{\centering Vanilla Prompting}
   &  1 & \cellcolor[HTML]{8FBCE8}8.2 {\scriptsize [7.9, 8.5]} & \cellcolor[HTML]{8FBCE8}8.6 {\scriptsize [8.3, 8.8]} \\
&  &  5 & \cellcolor[HTML]{8FBCE8}9.6 {\scriptsize [9.4, 9.7]} & \cellcolor[HTML]{8FBCE8}9.6 {\scriptsize [9.5, 9.7]} \\
&  & 30 & \cellcolor[HTML]{8FBCE8}9.5 {\scriptsize [9.4, 9.7]} & \cellcolor[HTML]{8FBCE8}9.7 {\scriptsize [9.6, 9.8]} \\
\cmidrule{2-5}
& \multirow{3}{\linewidth}{\centering Taxonomy-Guided Prompting}
   &  1 & \cellcolor[HTML]{F58D8D}1.3 {\scriptsize [1.2, 1.4]} & \cellcolor[HTML]{F58D8D}1.3 {\scriptsize [1.2, 1.3]} \\
&  &  5 & \cellcolor[HTML]{F9C7C7}2.8 {\scriptsize [2.5, 3.0]} & \cellcolor[HTML]{F9C7C7}2.9 {\scriptsize [2.6, 3.1]} \\
&  & 30 & \cellcolor[HTML]{8FBCE8}9.3 {\scriptsize [9.1, 9.5]} & \cellcolor[HTML]{8FBCE8}9.4 {\scriptsize [9.3, 9.6]} \\
\cmidrule{2-5}
& \multirow{3}{\linewidth}{\centering Supervised Fine-tuning}
   &  1 & \cellcolor[HTML]{CBE8B4}5.3 {\scriptsize [4.9, 5.7]} & \cellcolor[HTML]{CBE8B4}5.7 {\scriptsize [5.4, 6.1]} \\
&  &  5 & \cellcolor[HTML]{B6D5F5}6.6 {\scriptsize [6.3, 7.0]} & \cellcolor[HTML]{B6D5F5}7.2 {\scriptsize [6.8, 7.5]} \\
&  & 30 & \cellcolor[HTML]{B6D5F5}6.3 {\scriptsize [5.9, 6.7]} & \cellcolor[HTML]{B6D5F5}6.7 {\scriptsize [6.3, 7.0]} \\
\midrule
\multirow{7}{*}{\rotatebox{90}{GPT-5.1}}
& \multirow{3}{\linewidth}{\centering Vanilla Prompting}
   &  1 & \cellcolor[HTML]{8FBCE8}9.2 {\scriptsize [9.0, 9.4]} & \cellcolor[HTML]{8FBCE8}9.2 {\scriptsize [8.9, 9.4]} \\
&  &  5 & \cellcolor[HTML]{8FBCE8}9.7 {\scriptsize [9.6, 9.8]} & \cellcolor[HTML]{8FBCE8}9.8 {\scriptsize [9.7, 9.9]} \\
&  & 30 & \cellcolor[HTML]{8FBCE8}9.8 {\scriptsize [9.7, 9.9]} & \cellcolor[HTML]{8FBCE8}9.9 {\scriptsize [9.8, 9.9]} \\
\cmidrule{2-5}
& \multirow{3}{\linewidth}{\centering Taxonomy-Guided Prompting}
   &  1 & \cellcolor[HTML]{CBE8B4}4.6 {\scriptsize [4.2, 5.0]} & \cellcolor[HTML]{CBE8B4}4.5 {\scriptsize [4.2, 4.9]} \\
&  &  5 & \cellcolor[HTML]{B6D5F5}6.5 {\scriptsize [6.1, 6.9]} & \cellcolor[HTML]{B6D5F5}6.5 {\scriptsize [6.2, 6.9]} \\
&  & 30 & \cellcolor[HTML]{8FBCE8}9.3 {\scriptsize [9.1, 9.5]} & \cellcolor[HTML]{8FBCE8}9.3 {\scriptsize [9.1, 9.5]} \\
\midrule
\multirow{5}{*}{\rotatebox{90}{\scalebox{0.85}[1]{Claude Opus 4}}}
& \multirow{2}{\linewidth}{\centering Vanilla Prompting}
   &  1 & \cellcolor[HTML]{8FBCE8}9.8 {\scriptsize [9.7, 9.9]} & \cellcolor[HTML]{8FBCE8}9.9 {\scriptsize [9.8, 9.9]} \\
&  &  5 & \cellcolor[HTML]{8FBCE8}9.8 {\scriptsize [9.7, 9.9]} & \cellcolor[HTML]{8FBCE8}9.9 {\scriptsize [9.8, 9.9]} \\
\cmidrule{2-5}
& \multirow{2}{\linewidth}{\centering Taxonomy-Guided Prompting}
   &  1 & \cellcolor[HTML]{F58D8D}1.5 {\scriptsize [1.4, 1.6]} & \cellcolor[HTML]{F58D8D}1.4 {\scriptsize [1.3, 1.5]} \\
&  &  5 & \cellcolor[HTML]{F58D8D}1.7 {\scriptsize [1.6, 1.9]} & \cellcolor[HTML]{F58D8D}1.6 {\scriptsize [1.5, 1.7]} \\
\bottomrule
\end{tabular}
\caption{Results of \cocoeval's overall evaluation using conversation-level scores on a 1--10~scale, evaluated by o4-mini. Each setup is evaluated on 300~simulated conversations for \cocoevalbench. The intervals represent 95\%~bootstrap confidence intervals.}
\label{tab:overall-evaluation}
\end{table}

\subsection{Human Conversations} \label{sec:results-human-conversations}

The overall evaluation results in Table~\ref{tab:overall-evaluation} show consistency and collaborativeness scores of 5.2 and 5.7 out of 10 for human conversations, indicating that they are not perfectly consistent or collaborative. The fine-grained evaluation results in Figure~\ref{fig:fine-grained-human} provide a more detailed view, showing that human conversations contain all 10 evaluated behavior types at moderate frequencies, with Unclear Intent, Under Answering, Off Topic, and Repetition occurring relatively frequently.

\begin{takeawaybox}{Answer to RQ1 about Human Conversations}
Human conversations involve diverse types of inconsistent and uncollaborative behaviors.
\end{takeawaybox}

\subsection{Vanilla Prompting} \label{sec:results-vanilla-prompting}

\begin{figure}[t!]
    \centering
\begin{subfigure}{\linewidth}
    \includegraphics[width=\linewidth,trim={.3cm 3.8cm 0 0},clip]{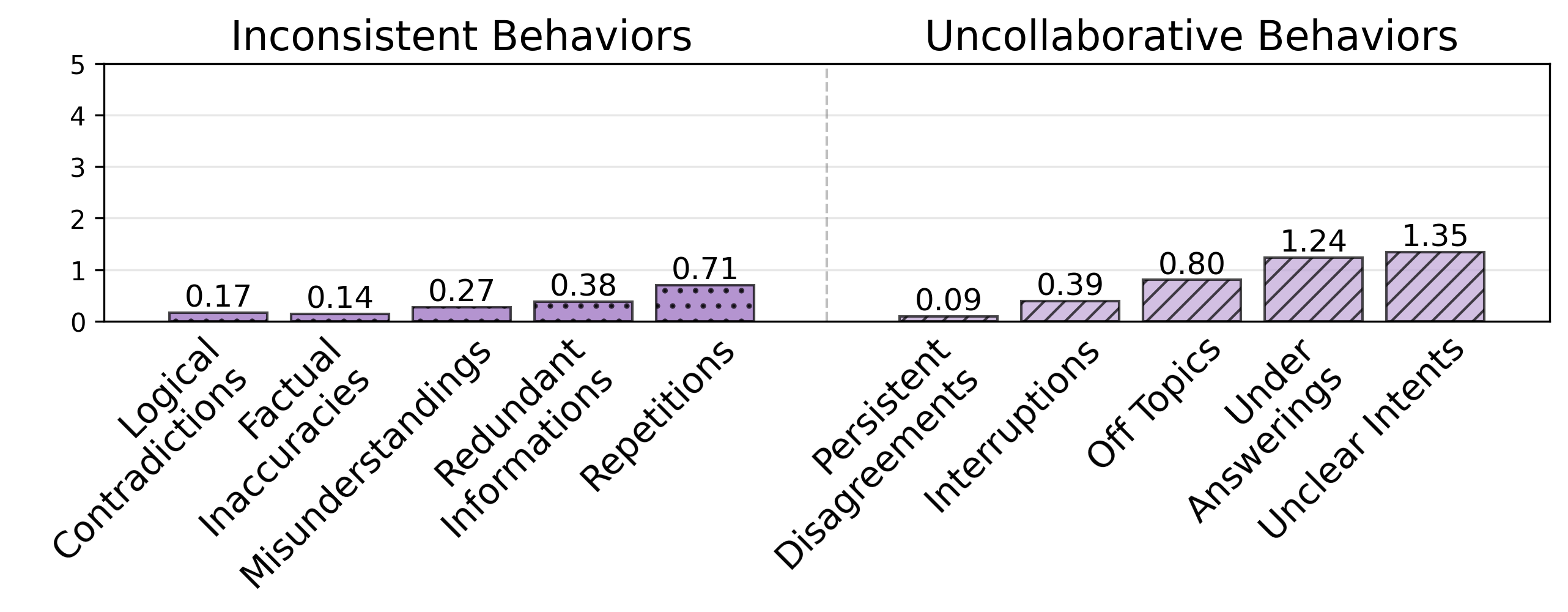}
    \caption{Human Conversations} \label{fig:fine-grained-human}
\end{subfigure}
\vskip .35em
\begin{subfigure}{\linewidth}
    \includegraphics[width=\linewidth,trim={.3cm 3.8cm 0 .75cm},clip]{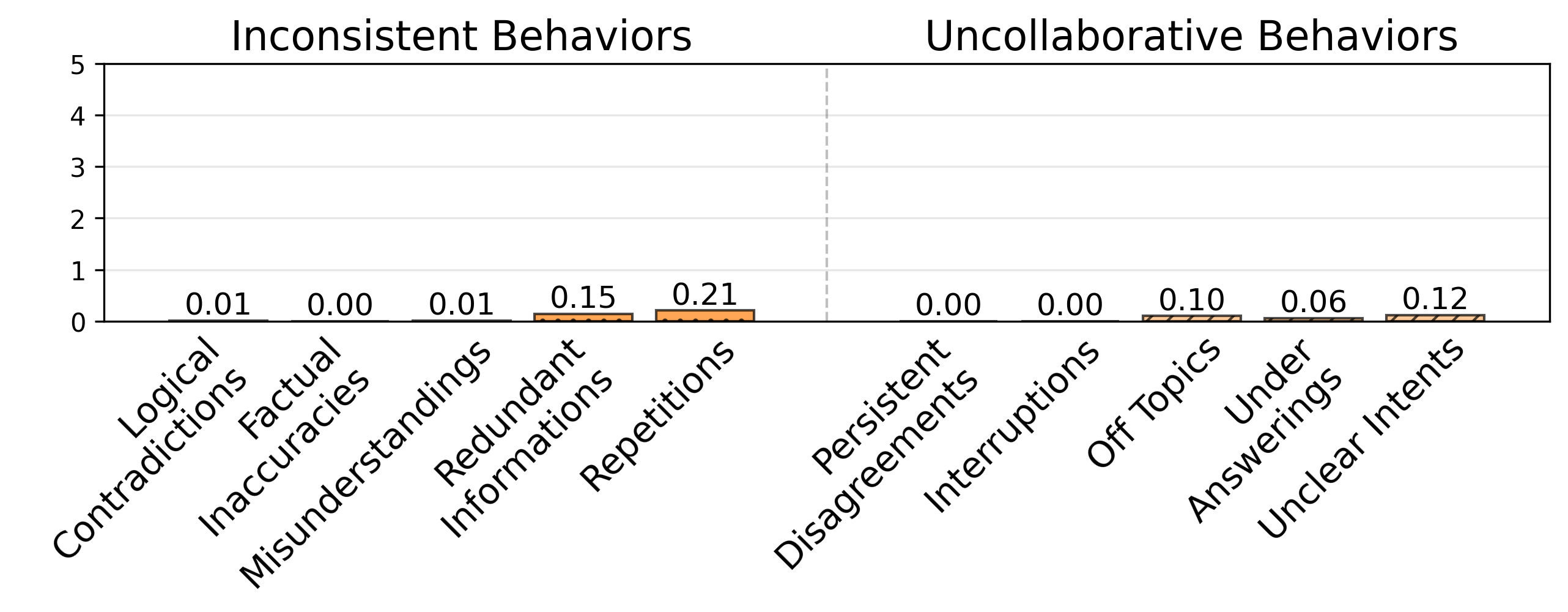}
    \caption{Vanilla prompting with 30 turns generated per LLM call} \label{fig:vanilla-30turn}
\end{subfigure}
\vskip .35em
\begin{subfigure}{\linewidth}
    \includegraphics[width=\linewidth,trim={.3cm 3.8cm 0 .75cm},clip]{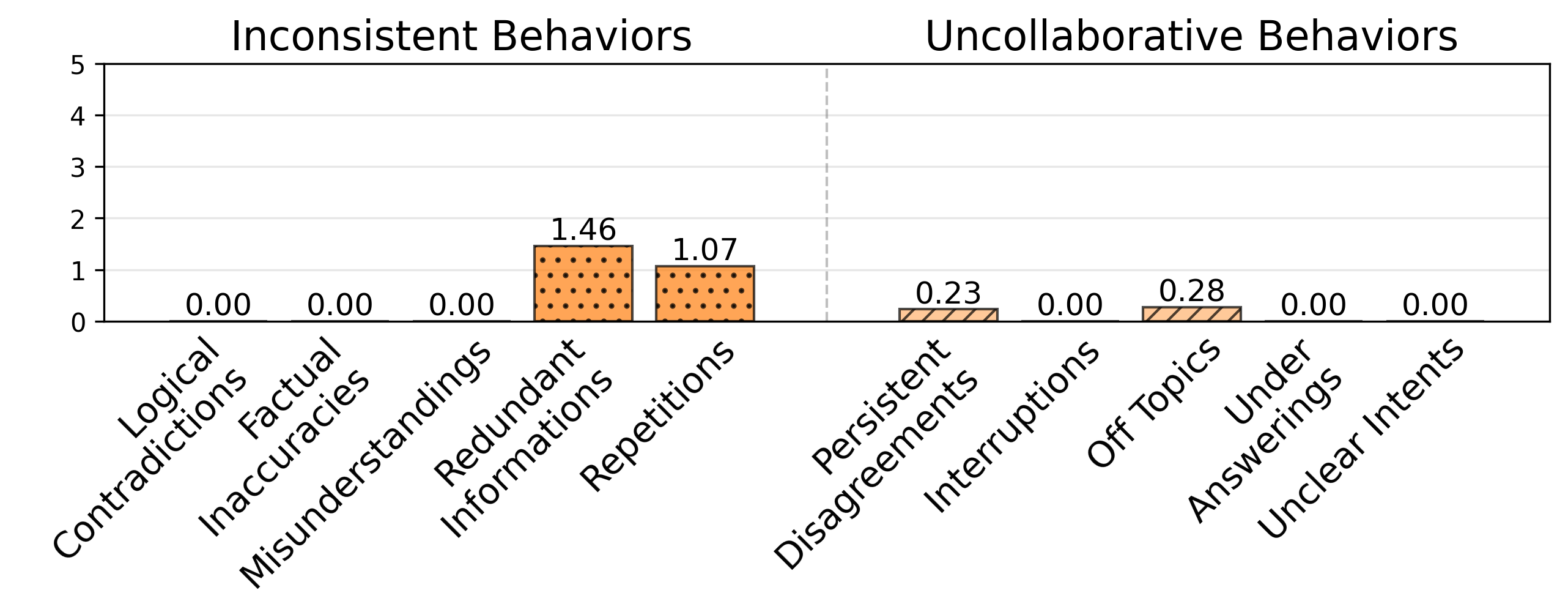}
    \caption{Vanilla prompting with 1 turn generated per LLM call} \label{fig:vanilla-1turn}
\end{subfigure}
\vskip .35em
\begin{subfigure}{\linewidth}
    \includegraphics[width=\linewidth,trim={.3cm 3.8cm 0 .75cm},clip]{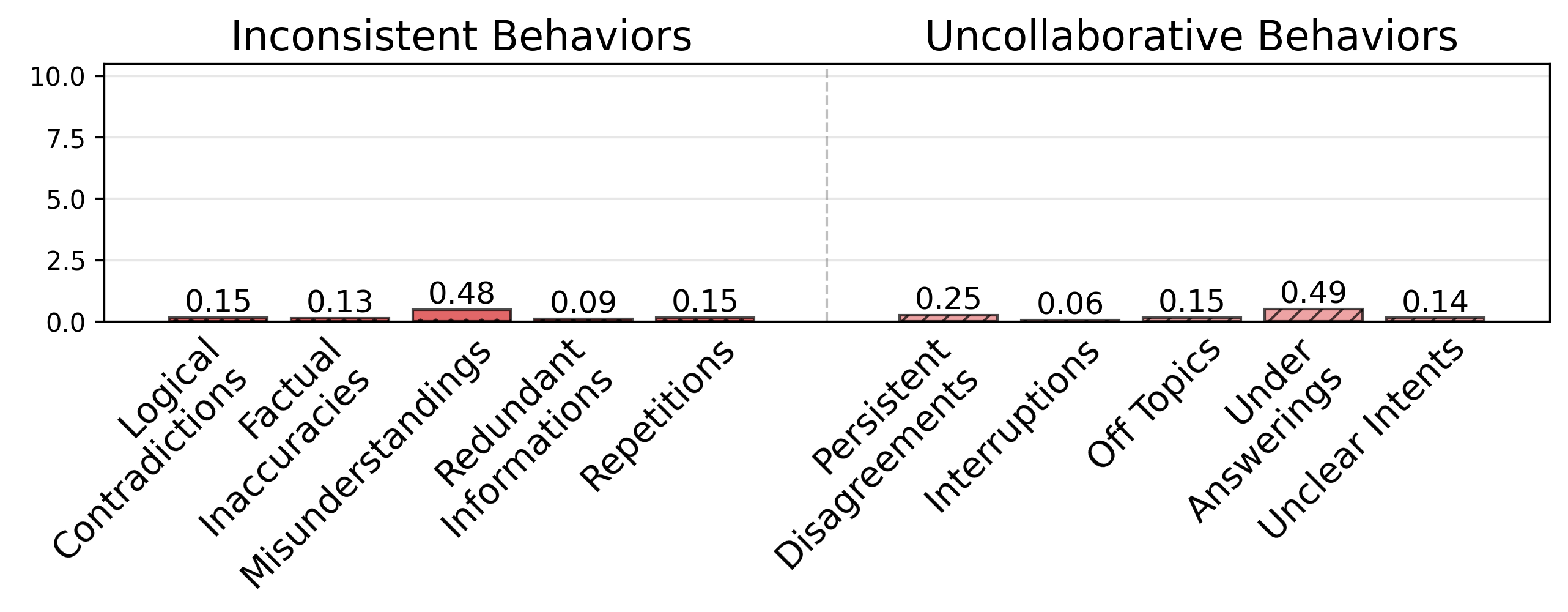}
    \caption{Taxonomy-guided prompting with 30 turns per call} \label{fig:taxsonomy-guided-30turns}
\end{subfigure}
\vskip .35em
\begin{subfigure}{\linewidth}
    \includegraphics[width=\linewidth,trim={.3cm 3.8cm 0 .75cm},clip]{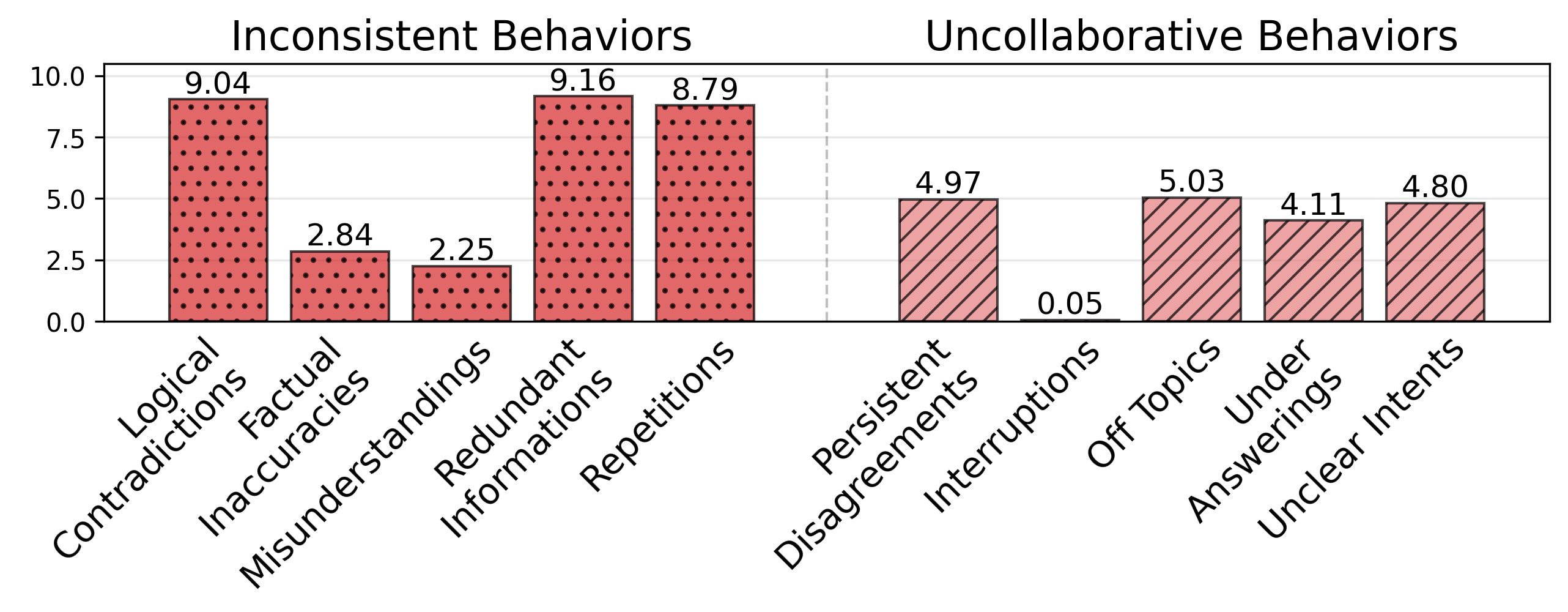}
    \caption{Taxonomy-guided prompting with 1 turn per call} \label{fig:taxsonomy-guided-1turn}
\end{subfigure}
\vskip .35em
\begin{subfigure}{\linewidth}
    \includegraphics[width=\linewidth,trim={.3cm 3.8cm 0 .75cm},clip]{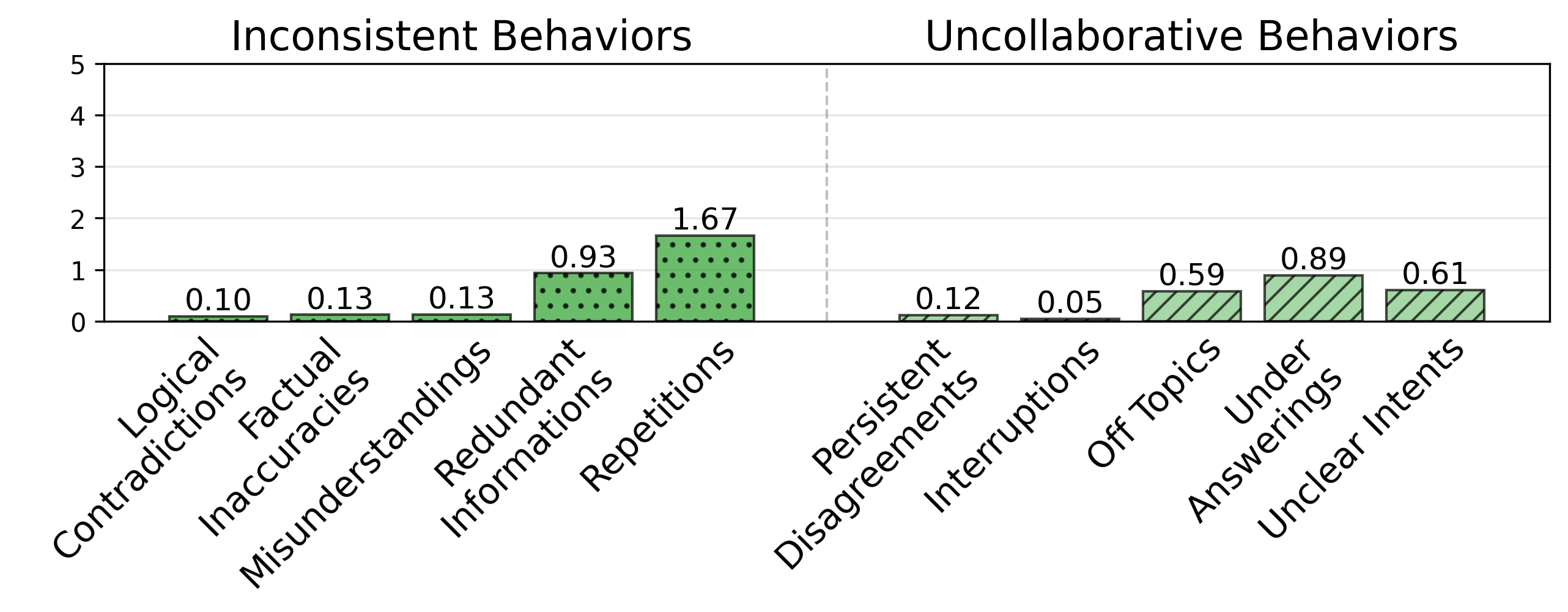}
    \caption{SFT with 30 turns generated per LLM call}
\end{subfigure}
\vskip .35em
\begin{subfigure}{\linewidth}
    \includegraphics[width=\linewidth,trim={.3cm .3cm 0 .75cm},clip]{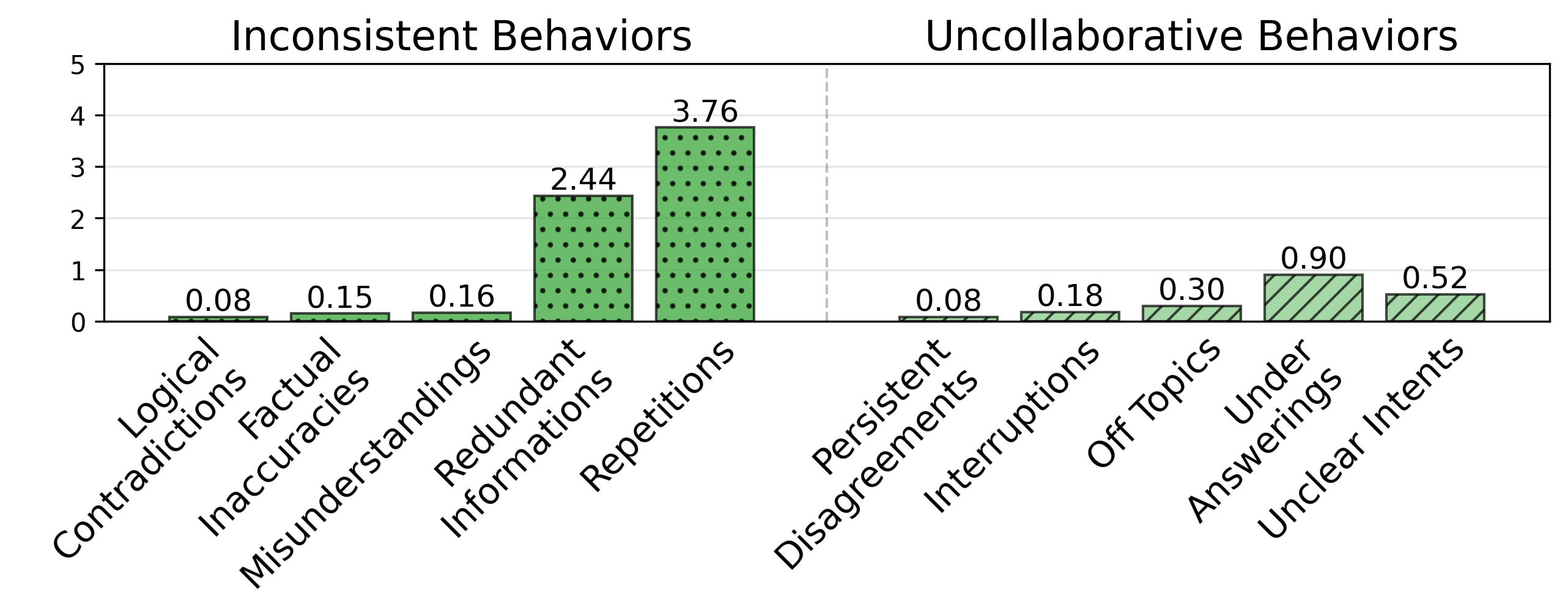}
    \caption{SFT with 1 turn generated per LLM call} \label{fig:sft-1turn}
\end{subfigure}
    \caption{Results of \cocoeval's fine-grained evaluation showing the number of turns exhibiting each behavior type per 30-turn conversation continuation by humans and GPT-4.1 simulation, averaged over 300 conversations in \cocoevalbench. The y-axes for taxonomy-guided prompting use different scales.}
    \label{fig:bargraph}
\end{figure}

The overall evaluation in Table~\ref{tab:overall-evaluation} indicates that conversations generated with vanilla prompting exhibit substantially higher consistency and collaborativeness than human conversations. The fine-grained evaluation in Figures~\ref{fig:vanilla-30turn} and \ref{fig:vanilla-1turn} further shows that inconsistent and uncollaborative behaviors are nearly absent, particularly when 30~turns are generated per call, highlighting substantial differences from human conversations~(Figure~\ref{fig:fine-grained-human}).

\begin{takeawaybox}{Answer to RQ2 about Vanilla Prompting}
With vanilla prompting, LLM-simulated conversations exhibit far fewer inconsistent and uncollaborative behaviors than human conversations.
\end{takeawaybox}

\subsection{Prompt Engineering} \label{sec:results-prompt-engineering}

The overall evaluation in Table~\ref{tab:overall-evaluation} shows that taxonomy-guided prompting substantially reduces overall consistency and collaborativeness relative to vanilla prompting. The effect is stronger when fewer turns are generated within a single LLM call. The fine-grained evaluation in Figure~\ref{fig:taxsonomy-guided-1turn} shows that this prompt substantially increases almost all types of inconsistent and uncollaborative behaviors when generating a single turn per call, although this prompt does not instruct the model to increase these behaviors but rather to reproduce their frequencies in human conversations. The resulting frequencies differ substantially from those observed in human conversations~(Figure~\ref{fig:fine-grained-human}).

\begin{takeawaybox}{Answer to RQ3 about Prompt Engineering}
Prompt engineering does not provide reliable control over inconsistent and uncollaborative behaviors and can lead to their overproduction when explicit instructions are provided.
\end{takeawaybox}

\subsection{SFT on Human Conversations} \label{sec:results-finetuning}

The overall evaluation in Table~\ref{tab:overall-evaluation} shows that GPT-4.1 fine-tuned on human conversations with SFT exhibits levels of consistency and collaborativeness comparable to those of human conversations, especially when generating one turn per call. However, the fine-grained evaluation in Figure~\ref{fig:sft-1turn} shows that the frequencies of inconsistent and uncollaborative behaviors do not match those of human conversations~(Figure~\ref{fig:fine-grained-human}), with fine-tuned GPT-4.1 overproducing Repetition and Redundant Information. We attempt to alleviate this issue using a presence penalty~(Appendix~\ref{appendix:repetition-penalty}), but we observe that the bias toward repetition and redundancy is strong and cannot be easily mitigated. These results indicate that even when fine-tuned on human conversations, LLM-simulated conversations still deviate from human conversations.

\begin{takeawaybox}{Answer to RQ4 about SFT}
SFT on human conversations can lead LLMs to overproduce a narrow subset of inconsistent behaviors, such as repetitions and redundancy, remaining misaligned with human conversations.
\end{takeawaybox}

\subsection{Generating Multiple Turns Per LLM Call} \label{sec:results-multiple-turns}

The results in Table~\ref{tab:overall-evaluation} and Figure~\ref{fig:bargraph} consistently show that generating multiple turns per LLM call, particularly 30~turns, introduces a strong bias toward reducing inconsistent and uncollaborative behaviors. In particular, Figure~\ref{fig:taxsonomy-guided-30turns} shows that generating 30 turns in a single LLM call almost eliminates the effect of taxonomy-guided prompting, whereas it substantially increases these behaviors when one turn is generated per call. However, generating fewer turns per call still fails to reproduce the distribution of these behaviors observed in human conversations~(Figure~\ref{fig:fine-grained-human}) and instead tends to overrepresent repetitive and redundant utterances, even with vanilla prompting, as shown in Figure~\ref{fig:vanilla-1turn}.

\begin{takeawaybox}{Answer to RQ5 about Generation Strategy}
The number of turns generated in a single LLM call strongly influences inconsistent and uncollaborative behaviors in simulated conversations, but adjusting this parameter alone does not fully account for the gap from human conversations.
\end{takeawaybox}

\section{Benefits of \cocoeval's Evaluation} \label{sec:discussion}

The results in Section~\ref{sec:results} show that \cocoeval's fine-grained evaluation provides a detailed and interpretable evaluation, and as a result, it captures differences between human and LLM-simulated conversations that are not captured by conventional conversation-level Likert-scale evaluation~(the overall evaluation).
First, the overall evaluation in Table~\ref{tab:overall-evaluation} assigns similar scores to human conversations and LLM-generated conversations for fine-tuned GPT-4.1 when generating one turn per call. However, our manual analysis shows that conversations simulated by fine-tuned GPT-4.1 often exhibit unnatural repetition, as shown in an example in Table~\ref{tab:finetuned-example-repetition}, where the same speaker repeatedly expresses agreement with prior utterances. In contrast, the fine-grained evaluation in Figure~\ref{fig:sft-1turn} successfully identifies that fine-tuned GPT-4.1 overproduces repetition and redundancy compared to human conversations~(Figure~\ref{fig:fine-grained-human}).
In addition, GPT-5.1 with taxonomy-guided prompting~(generating one turn per call) achieves scores similar to those of human conversations in the overall evaluation in Table~\ref{tab:overall-evaluation}, while our manual analysis shows that these conversations often contain unnaturally frequent misunderstandings, as illustrated in Table~\ref{tab:gpt5-1-example-misunderstanding}. The fine-grained evaluation in Figure~\ref{fig:gpt-5.1-taxonomy-guided-1-turn} successfully captures this difference, revealing a substantially different distribution of turn-level behaviors, particularly a high frequency of misunderstanding.

\begin{table}[t]
\centering
\scriptsize
\begin{tabular}{M{.14\linewidth}L{.76\linewidth}}
\toprule
Manager   & Exactly. \\
\midrule
Interface & Yeah, that makes sense. \\
\midrule
Designer  & That's right. \\
\midrule
Interface & Alright. \\
\midrule
Manager   & Okay. \\
\bottomrule
\end{tabular}
\caption{Conversation generated by GPT-4.1 fine-tuned on human conversations for QMSum~Product, exhibiting unnatural repetitive utterances that are not human-like.}
\label{tab:finetuned-example-repetition}
\end{table}

\begin{table}[t]
\centering
\scriptsize
\begin{tabular}{M{.14\linewidth}L{.76\linewidth}}
\toprule
PhD C   & Mm-hmm, so you're saying we’d basically scale the acoustic posteriors with the VAD silence likelihood, right? [...] \\
\midrule
Prof. B & No, I wasn’t really talking about changing the posteriors [...] \\
\midrule
PhD A   & So we’d be tweaking the silence posteriors a bit? [...] \\
\midrule
Prof. B & No, you wouldn’t be touching the posteriors [...] \\
\bottomrule
\end{tabular}
\caption{Conversation generated by GPT-5.1 with taxonomy-guided prompting for QMSum~Academic, exhibiting unnaturally repetitive misunderstanding.}
\label{tab:gpt5-1-example-misunderstanding}
\end{table}

\section{Conclusion}

This is the first work to conduct a detailed analysis of LLM-simulated conversations with a focus on inconsistent and uncollaborative behaviors, which are inherent to human conversations and contribute to the complexity of human social interactions. We introduce \cocoeval{}, an evaluation framework for detailed comparisons between human and simulated conversations with respect to inconsistent and uncollaborative behaviors. Our analysis reveals substantial discrepancies between human and simulated conversations, raising concerns about the use of LLMs as proxies for human social interaction.

\section*{Limitations}

Our analysis is constrained by the limited availability of publicly accessible resources containing real-world multi-party human conversations. Existing public resources are heavily skewed toward governmental meetings and debates, while conversations from other domains such as business meetings remain scarce. Collecting conversations from a broader range of real-world settings would support the study of social interaction simulation across more diverse scenarios. Our study also focuses on conversations in English.

\ifreview\else
\section*{Acknowledgment}
This work was supported by NSF IIS-2338418. We thank Ngoc-Hieu Nguyen for his contributions as an annotator and for his valuable feedback.
\fi

\bibliography{custom}

@inproceedings{zhou-etal-2024-real,
    title = "Is this the real life? Is this just fantasy? The Misleading Success of Simulating Social Interactions With {LLM}s",
    author = "Zhou, Xuhui  and
      Su, Zhe  and
      Eisape, Tiwalayo  and
      Kim, Hyunwoo  and
      Sap, Maarten",
    editor = "Al-Onaizan, Yaser  and
      Bansal, Mohit  and
      Chen, Yun-Nung",
    booktitle = "Proceedings of the 2024 Conference on Empirical Methods in Natural Language Processing",
    month = nov,
    year = "2024",
    address = "Miami, Florida, USA",
    publisher = "Association for Computational Linguistics",
    url = "https://aclanthology.org/2024.emnlp-main.1208/",
    doi = "10.18653/v1/2024.emnlp-main.1208",
    pages = "21692--21714",
}

@inproceedings{
zhou2024sotopia,
title={{SOTOPIA}: Interactive Evaluation for Social Intelligence in Language Agents},
author={Xuhui Zhou and Hao Zhu and Leena Mathur and Ruohong Zhang and Haofei Yu and Zhengyang Qi and Louis-Philippe Morency and Yonatan Bisk and Daniel Fried and Graham Neubig and Maarten Sap},
booktitle={The Twelfth International Conference on Learning Representations},
year={2024},
url={https://openreview.net/forum?id=mM7VurbA4r}
}

@inproceedings{higashinaka-etal-2021-integrated,
    title = "Integrated taxonomy of errors in chat-oriented dialogue systems",
    author = "Higashinaka, Ryuichiro  and
      Araki, Masahiro  and
      Tsukahara, Hiroshi  and
      Mizukami, Masahiro",
    editor = "Li, Haizhou  and
      Levow, Gina-Anne  and
      Yu, Zhou  and
      Gupta, Chitralekha  and
      Sisman, Berrak  and
      Cai, Siqi  and
      Vandyke, David  and
      Dethlefs, Nina  and
      Wu, Yan  and
      Li, Junyi Jessy",
    booktitle = "Proceedings of the 22nd Annual Meeting of the Special Interest Group on Discourse and Dialogue",
    month = jul,
    year = "2021",
    address = "Singapore and Online",
    publisher = "Association for Computational Linguistics",
    url = "https://aclanthology.org/2021.sigdial-1.10/",
    doi = "10.18653/v1/2021.sigdial-1.10",
    pages = "89--98",
}

@inproceedings{zhong-etal-2021-qmsum,
    title = "{QMS}um: A New Benchmark for Query-based Multi-domain Meeting Summarization",
    author = "Zhong, Ming  and
      Yin, Da  and
      Yu, Tao  and
      Zaidi, Ahmad  and
      Mutuma, Mutethia  and
      Jha, Rahul  and
      Awadallah, Ahmed Hassan  and
      Celikyilmaz, Asli  and
      Liu, Yang  and
      Qiu, Xipeng  and
      Radev, Dragomir",
    editor = "Toutanova, Kristina  and
      Rumshisky, Anna  and
      Zettlemoyer, Luke  and
      Hakkani-Tur, Dilek  and
      Beltagy, Iz  and
      Bethard, Steven  and
      Cotterell, Ryan  and
      Chakraborty, Tanmoy  and
      Zhou, Yichao",
    booktitle = "Proceedings of the 2021 Conference of the North American Chapter of the Association for Computational Linguistics: Human Language Technologies",
    month = jun,
    year = "2021",
    address = "Online",
    publisher = "Association for Computational Linguistics",
    url = "https://aclanthology.org/2021.naacl-main.472/",
    doi = "10.18653/v1/2021.naacl-main.472",
    pages = "5905--5921",
}

@misc{ncpc2024,
    title = "{NCPC} Commission Meeting Archive",
    author = "NCPC",
    year = "2024",
    url = "https://www.ncpc.gov/review/archive/2024/index/",
}

@inproceedings{
fan2024topicconversation,
title={Topic-Conversation Relevance ({TCR})  Dataset and Benchmarks},
author={Yaran Fan and Jamie Pool and Senja Filipi and Ross Cutler},
booktitle={The Thirty-eight Conference on Neural Information Processing Systems Datasets and Benchmarks Track},
year={2024},
url={https://openreview.net/forum?id=qMuzlVmiQh}
}

@inproceedings{zhang-etal-2016-conversational,
    title = "Conversational Flow in {O}xford-style Debates",
    author = "Zhang, Justine  and
      Kumar, Ravi  and
      Ravi, Sujith  and
      Danescu-Niculescu-Mizil, Cristian",
    editor = "Knight, Kevin  and
      Nenkova, Ani  and
      Rambow, Owen",
    booktitle = "Proceedings of the 2016 Conference of the North {A}merican Chapter of the Association for Computational Linguistics: Human Language Technologies",
    month = jun,
    year = "2016",
    address = "San Diego, California",
    publisher = "Association for Computational Linguistics",
    url = "https://aclanthology.org/N16-1017/",
    doi = "10.18653/v1/N16-1017",
    pages = "136--141"
}

@misc{o3-and-o4-mini,
    title = "Introducing {O}pen{AI} o3 and o4-mini",
    author = "OpenAI",
    year = "2025",
    url = "https://openai.com/index/introducing-o3-and-o4-mini/",
}

@misc{gpt-4-1,
    title = "Introducing {GPT}-4.1 in the {API}",
    author = "OpenAI",
    year = "2025",
    url = "https://openai.com/index/gpt-4-1/",
}

@misc{gpt-5-1,
    title = "{GPT}-5.1: A smarter, more conversational {C}hat{GPT}",
    author = "OpenAI",
    year = "2025",
    url = "https://openai.com/index/gpt-5-1/",
}

@inproceedings{park2023generative,
author = {Park, Joon Sung and O'Brien, Joseph and Cai, Carrie Jun and Morris, Meredith Ringel and Liang, Percy and Bernstein, Michael S.},
title = {Generative Agents: Interactive Simulacra of Human Behavior},
year = {2023},
isbn = {9798400701320},
publisher = {Association for Computing Machinery},
address = {New York, NY, USA},
url = {https://doi.org/10.1145/3586183.3606763},
doi = {10.1145/3586183.3606763},
booktitle = {Proceedings of the 36th Annual ACM Symposium on User Interface Software and Technology},
articleno = {2},
numpages = {22},
location = {San Francisco, CA, USA},
series = {UIST '23}
}

@inproceedings{liu2025promediate,
    title = "{P}ro{M}ediate: A Simulation Testbed for Evaluating Proactive Mediation in Multi-Party Negotiation",
    author = "Liu, Ziyi  and
      Sarrafzadeh, Bahareh  and
      Zhou, Pei  and
      Yang, Longqi  and
      Zhao, Jieyu  and
      Sharma, Ashish",
    editor = "Liakata, Maria  and
      Moreira, Viviane P.  and
      Zhang, Jiajun  and
      Jurgens, David",
    booktitle = "Findings of the {A}ssociation for {C}omputational {L}inguistics: {ACL} 2026",
    month = jul,
    year = "2026",
    address = "San Diego, California, United States",
    publisher = "Association for Computational Linguistics",
    url = "https://aclanthology.org/2026.findings-acl.1479/",
    doi = "10.18653/v1/2026.findings-acl.1479",
    pages = "29570--29598",
    ISBN = "979-8-89176-395-1",
}

@inproceedings{
abdulhai2025consistently,
title={Consistently Simulating Human Personas with Multi-Turn Reinforcement Learning},
author={Marwa Abdulhai and Ryan Cheng and Donovan Clay and Tim Althoff and Sergey Levine and Natasha Jaques},
booktitle={The Thirty-ninth Annual Conference on Neural Information Processing Systems},
year={2025},
url={https://openreview.net/forum?id=A0T3piHiis}
}

@inproceedings{frisch-giulianelli-2024-llm,
    title = "{LLM} Agents in Interaction: Measuring Personality Consistency and Linguistic Alignment in Interacting Populations of Large Language Models",
    author = "Frisch, Ivar  and
      Giulianelli, Mario",
    editor = "Deshpande, Ameet  and
      Hwang, EunJeong  and
      Murahari, Vishvak  and
      Park, Joon Sung  and
      Yang, Diyi  and
      Sabharwal, Ashish  and
      Narasimhan, Karthik  and
      Kalyan, Ashwin",
    booktitle = "Proceedings of the 1st Workshop on Personalization of Generative AI Systems (PERSONALIZE 2024)",
    month = mar,
    year = "2024",
    address = "St. Julians, Malta",
    publisher = "Association for Computational Linguistics",
    url = "https://aclanthology.org/2024.personalize-1.9/",
    pages = "102--111",
}

@inproceedings{bianchi2024how,
author = {Bianchi, Federico and Chia, Patrick John and Yuksekgonul, Mert and Tagliabue, Jacopo and Jurafsky, Dan and Zou, James},
title = {How well can LLMs negotiate? NEGOTIATIONARENA platform and analysis},
year = {2024},
publisher = {JMLR},
booktitle = {Proceedings of the 41st International Conference on Machine Learning},
articleno = {158},
numpages = {17},
location = {Vienna, Austria},
}

@inproceedings{
piedrahita2025corrupted,
title={Corrupted by Reasoning: Reasoning Language Models Become Free-Riders in Public Goods Games},
author={David Guzman Piedrahita and Yongjin Yang and Mrinmaya Sachan and Giorgia Ramponi and Bernhard Sch{\"o}lkopf and Zhijing Jin},
booktitle={Second Conference on Language Modeling},
year={2025},
url={https://openreview.net/forum?id=kH6LOHGjEl}
}

@inproceedings{
piatti2024cooperate,
title={Cooperate or Collapse:  Emergence of Sustainable Cooperation in a Society of {LLM} Agents},
author={Giorgio Piatti and Zhijing Jin and Max Kleiman-Weiner and Bernhard Sch{\"o}lkopf and Mrinmaya Sachan and Rada Mihalcea},
booktitle={The Thirty-eighth Annual Conference on Neural Information Processing Systems},
year={2024},
url={https://openreview.net/forum?id=0zWzJj6lO3}
}

@inproceedings{maharana-etal-2024-evaluating,
    title = "Evaluating Very Long-Term Conversational Memory of {LLM} Agents",
    author = "Maharana, Adyasha  and
      Lee, Dong-Ho  and
      Tulyakov, Sergey  and
      Bansal, Mohit  and
      Barbieri, Francesco  and
      Fang, Yuwei",
    editor = "Ku, Lun-Wei  and
      Martins, Andre  and
      Srikumar, Vivek",
    booktitle = "Proceedings of the 62nd Annual Meeting of the Association for Computational Linguistics (Volume 1: Long Papers)",
    month = aug,
    year = "2024",
    address = "Bangkok, Thailand",
    publisher = "Association for Computational Linguistics",
    url = "https://aclanthology.org/2024.acl-long.747/",
    doi = "10.18653/v1/2024.acl-long.747",
    pages = "13851--13870",
}

@INPROCEEDINGS{wang2025multiparty,
  author={Wang, Xiaoyu and Xi, Ningyuan and Gu, Qingqing and Ji, Luo},
  booktitle={2025 International Joint Conference on Neural Networks (IJCNN)}, 
  title={Multi-Party Supervised Fine-tuning of Language Models for Multi-Party Dialogue Generation}, 
  year={2025},
  volume={},
  number={},
  pages={1-9},
  doi={10.1109/IJCNN64981.2025.11228310}}

@inproceedings{pang-etal-2020-towards,
    title = "Towards Holistic and Automatic Evaluation of Open-Domain Dialogue Generation",
    author = "Pang, Bo  and
      Nijkamp, Erik  and
      Han, Wenjuan  and
      Zhou, Linqi  and
      Liu, Yixian  and
      Tu, Kewei",
    editor = "Jurafsky, Dan  and
      Chai, Joyce  and
      Schluter, Natalie  and
      Tetreault, Joel",
    booktitle = "Proceedings of the 58th Annual Meeting of the Association for Computational Linguistics",
    month = jul,
    year = "2020",
    address = "Online",
    publisher = "Association for Computational Linguistics",
    url = "https://aclanthology.org/2020.acl-main.333/",
    doi = "10.18653/v1/2020.acl-main.333",
    pages = "3619--3629",
}

@inproceedings{li-etal-2023-normdial,
    title = "{N}orm{D}ial: A Comparable Bilingual Synthetic Dialog Dataset for Modeling Social Norm Adherence and Violation",
    author = "Li, Oliver  and
      Subramanian, Mallika  and
      Saakyan, Arkadiy  and
      CH-Wang, Sky  and
      Muresan, Smaranda",
    editor = "Bouamor, Houda  and
      Pino, Juan  and
      Bali, Kalika",
    booktitle = "Proceedings of the 2023 Conference on Empirical Methods in Natural Language Processing",
    month = dec,
    year = "2023",
    address = "Singapore",
    publisher = "Association for Computational Linguistics",
    url = "https://aclanthology.org/2023.emnlp-main.974/",
    doi = "10.18653/v1/2023.emnlp-main.974",
    pages = "15732--15744",
}

@article{sreedhar2024simulating,
      title={Simulating Human Strategic Behavior: Comparing Single and Multi-agent LLMs}, 
      author={Karthik Sreedhar and Lydia Chilton},
      year={2024},
      journal={arXiv preprint arXiv:2402.08189},
}

@article{pickering2004toward,
title={Toward a mechanistic psychology of dialogue}, volume={27}, DOI={10.1017/S0140525X04000056}, number={2}, journal={Behavioral and Brain Sciences}, author={Pickering, Martin J. and Garrod, Simon}, year={2004}, pages={169–190}
}

@inproceedings{mahajan-etal-2022-towards,
    title = "Towards Evaluation of Multi-party Dialogue Systems",
    author = "Mahajan, Khyati  and
      Santhanam, Sashank  and
      Shaikh, Samira",
    editor = "Shaikh, Samira  and
      Ferreira, Thiago  and
      Stent, Amanda",
    booktitle = "Proceedings of the 15th International Conference on Natural Language Generation",
    month = jul,
    year = "2022",
    address = "Waterville, Maine, USA and virtual meeting",
    publisher = "Association for Computational Linguistics",
    url = "https://aclanthology.org/2022.inlg-main.23/",
    doi = "10.18653/v1/2022.inlg-main.23",
    pages = "278--287"
}

@inproceedings{
chu2024cohesive,
title={Cohesive Conversations: Enhancing Authenticity in Multi-Agent Simulated Dialogues},
author={KuanChao Chu and Yi-Pei Chen and Hideki Nakayama},
booktitle={First Conference on Language Modeling},
year={2024},
url={https://openreview.net/forum?id=3ypWPhMGhV}
}

@article{nonomura2025who,
  author  = {Nonomura, Ryota and Mori, Hiroki},
  title   = {Who speaks next? Multi-party AI discussion leveraging the systematics of turn-taking in Murder Mystery games},
  journal = {Frontiers in Artificial Intelligence},
  volume  = {8},
  pages   = {1582287},
  year    = {2025},
  doi     = {10.3389/frai.2025.1582287},
  url     = {https://doi.org/10.3389/frai.2025.1582287}
}

@article{schegloff1992repair,
 ISSN = {00029602, 15375390},
 URL = {http://www.jstor.org/stable/2781417},
 author = {Emanuel A. Schegloff},
 journal = {American Journal of Sociology},
 number = {5},
 pages = {1295--1345},
 publisher = {University of Chicago Press},
 title = {Repair After Next Turn: The Last Structurally Provided Defense of Intersubjectivity in Conversation},
 urldate = {2026-01-19},
 volume = {97},
 year = {1992}
}

@misc{claude4,
    author       = {Anthropic},
    title        = {System Card: {C}laude {O}pus 4 \& {C}laude {S}onnet 4},
    year         = {2025},
    url          = {https://www-cdn.anthropic.com/6d8a8055020700718b0c49369f60816ba2a7c285.pdf},
}

@inproceedings{
wu2025collabllm,
title={Collab{LLM}: From Passive Responders to Active Collaborators},
author={Shirley Wu and Michel Galley and Baolin Peng and Hao Cheng and Gavin Li and Yao Dou and Weixin Cai and James Zou and Jure Leskovec and Jianfeng Gao},
booktitle={Forty-second International Conference on Machine Learning},
year={2025},
url={https://openreview.net/forum?id=DmH4HHVb3y}
}

@inproceedings{wang-etal-2023-chatgpt-defend,
    title = "Can {C}hat{GPT} Defend its Belief in Truth? Evaluating {LLM} Reasoning via Debate",
    author = "Wang, Boshi  and
      Yue, Xiang  and
      Sun, Huan",
    editor = "Bouamor, Houda  and
      Pino, Juan  and
      Bali, Kalika",
    booktitle = "Findings of the Association for Computational Linguistics: EMNLP 2023",
    month = dec,
    year = "2023",
    address = "Singapore",
    publisher = "Association for Computational Linguistics",
    url = "https://aclanthology.org/2023.findings-emnlp.795/",
    doi = "10.18653/v1/2023.findings-emnlp.795",
    pages = "11865--11881",
}

@inproceedings{
chuang2025debate,
title={{DEBATE}: A Large-Scale Benchmark for Role-Playing {LLM} Agents in Multi-Agent, Long-Form Debates},
author={Yun-Shiuan Chuang and Ruixuan Tu and Chengtao Dai and Smit Vasani and Binwei Yao and Michael Henry Tessler and Sijia Yang and Dhavan V. Shah and Robert D. Hawkins and Junjie Hu and Timothy T. Rogers},
booktitle={Workshop on Scaling Environments for Agents},
year={2025},
url={https://openreview.net/forum?id=7mWVbd4IXD}
}

@article{elhilali2025large,
  title   = {Large Language Model–Based Patient Simulation to Foster Communication Skills in Health Care Professionals: User-Centered Development and Usability Study},
  author  = {Elhilali, Ahmed and Ngo, Andy Suy-Huor and Reichenpfader, Daniel and Denecke, Kerstin},
  journal = {JMIR Medical Education},
  year    = {2025},
  volume  = {11},
  pages   = {e81271},
  doi     = {10.2196/81271},
  date    = {2025-12-12}
}

@inproceedings{zheng-etal-2023-augesc,
    title = "{A}ug{ESC}: Dialogue Augmentation with Large Language Models for Emotional Support Conversation",
    author = "Zheng, Chujie  and
      Sabour, Sahand  and
      Wen, Jiaxin  and
      Zhang, Zheng  and
      Huang, Minlie",
    editor = "Rogers, Anna  and
      Boyd-Graber, Jordan  and
      Okazaki, Naoaki",
    booktitle = "Findings of the Association for Computational Linguistics: ACL 2023",
    month = jul,
    year = "2023",
    address = "Toronto, Canada",
    publisher = "Association for Computational Linguistics",
    url = "https://aclanthology.org/2023.findings-acl.99/",
    doi = "10.18653/v1/2023.findings-acl.99",
    pages = "1552--1568",
}

@techreport{manning2024automated,
 title = "Automated Social Science: Language Models as Scientist and Subjects",
 author = "Manning, Benjamin S and Zhu, Kehang and Horton, John J",
 institution = "National Bureau of Economic Research",
 type = "Working Paper",
 series = "Working Paper Series",
 number = "32381",
 year = "2024",
 month = "April",
 doi = {10.3386/w32381},
 URL = "http://www.nber.org/papers/w32381",
}

@incollection{grice1975logic,
  author    = {Grice, H. Paul},
  title     = {Logic and Conversation},
  booktitle = {Speech Acts},
  editor    = {Cole, Peter and Morgan, Jerry L.},
  series    = {Syntax and Semantics},
  volume    = {3},
  pages     = {41--58},
  year      = {1975},
  publisher = {Academic Press},
  address   = {New York},
  doi       = {10.1163/9789004368811_003}
}

@article{yan2025sharechat,
      title={ShareChat: A Dataset of Chatbot Conversations in the Wild}, 
      author={Yueru Yan and Tuc Nguyen and Bo Su and Melissa Lieffers and Thai Le},
      year={2026},
      journal={arXiv preprint arXiv:2512.17843}
}

\let\addcontentsline\oldaddcontentsline

\newpage
\setcounter{tocdepth}{2}
\renewcommand{\contentsname}{\centering Table of Contents of Appendix}
\tableofcontents

\appendix

\newpage

\section{\cocoevalbench}

This section provides additional details on \cocoevalbench{} explained in Section~\ref{sec:cocoevalbench}.

\subsection{Conversation Sources and Licenses}

\cocoevalbench{} is constructed from four public resources containing real-world human conversations. Table~\ref{tab:source-datasets} lists the URLs used to access these resources~(QMSum includes three subsets). QMSum and IQ2 are published under the MIT License, and NCPC and SIM are published under the CC BY-SA 4.0 License. \cocoevalbench{} is licensed under CC BY-SA 4.0, which is compatible with the licenses of all source datasets.

\begin{table}[t]
    \centering
    \small
    \begin{tabular}{lL{.75\linewidth}}
    \toprule
        Dataset & URL \\
    \midrule
        QMSum & \url{https://github.com/Yale-LILY/QMSum} \\
    \midrule
        NCPC  & \url{https://github.com/microsoft/topic_conversation} \\
    \midrule
        SIM   & \url{https://github.com/microsoft/topic_conversation} \\
    \midrule
        IQ2   & \url{https://tisjune.github.io/datasets/iq2_data_release.zip} \\
    \bottomrule
    \end{tabular}
    \caption{Source datasets for \cocoevalbench.}
    \label{tab:source-datasets}
\end{table}

\subsection{Statistics}

Table~\ref{tab:statistics-detail} shows detailed statistics of the evaluation and training sets of \cocoevalbench.

\begin{table*}[t]
\centering
\scriptsize
\setlength{\tabcolsep}{5pt}
\begin{subtable}{\linewidth}
\centering
\begin{tabular}{cccrrrrrrrrrrrr}
\toprule
\multirow{2.7}{*}{Category} & \multirow{2.7}{*}{Dataset} & \multirow{2.7}{*}{\# Data} & \multicolumn{6}{c}{\# participants} & \multicolumn{6}{c}{Sequence length} \\
\cmidrule(lr){4-9}\cmidrule(lr){10-15}
 & & & min & 25\% & med & 75\% & max & mean & min & 25\% & med & 75\% & max & mean \\
\midrule
Business
& QMSum Product & 50 & 4.0 & 4.0 & 4.0 & 4.0 & 4.0 & 4.0 & 1.0 & 2.0 & 5.0 & 9.0 & 289.0 & 8.0 \\
\midrule
Academic
& QMSum Academic & 50 & 4.0 & 5.0 & 6.0 & 7.0 & 7.0 & 5.7 & 1.0 & 1.0 & 5.0 & 12.0 & 291.0 & 10.0 \\
\midrule
\multirow{2}{*}{\centering Governmental}
& QMSum Committee & 50 & 7.0 & 9.0 & 11.0 & 11.0 & 51.0 & 18.5 & 1.0 & 10.0 & 40.5 & 83.0 & 756.0 & 66.3 \\
& NCPC & 50 & 17.0 & 17.0 & 28.0 & 34.0 & 34.0 & 25.8 & 1.0 & 3.0 & 16.0 & 108.0 & 150.0 & 50.4 \\
\midrule
\multirow{2}{*}{\centering Debates}
& SIM & 50 & 4.0 & 4.0 & 4.0 & 4.0 & 4.0 & 4.0 & 1.0 & 1.0 & 6.0 & 21.0 & 138.0 & 16.2 \\
& IQ2 & 50 & 7.0 & 8.0 & 9.0 & 12.0 & 16.0 & 10.1 & 1.0 & 4.0 & 13.0 & 48.0 & 1213.0 & 41.3 \\
\bottomrule
\end{tabular}
\caption{Evaluation Set} \label{tab:stat-eval-detail}
\end{subtable}
\vskip 1em
\begin{subtable}{\linewidth}
\centering
\begin{tabular}{cccrrrrrrrrrrrr}
\toprule
\multirow{2.7}{*}{Category} & \multirow{2.7}{*}{Dataset} & \multirow{2.7}{*}{\# Data} & \multicolumn{6}{c}{\# participants} & \multicolumn{6}{c}{Sequence length} \\
\cmidrule(lr){4-9}\cmidrule(lr){10-15}
 & & & min & 25\% & med & 75\% & max & mean & min & 25\% & med & 75\% & max & mean \\
\midrule
Business
& QMSum Product & 500 & 4.0 & 4.0 & 4.0 & 4.0 & 4.0 & 4.0 & 1.0 & 2.0 & 5.0 & 10.0 & 468.0 & 9.3 \\
\midrule
Governmental
& QMSum Committee & 300 & 6.0 & 8.0 & 10.0 & 11.0 & 64.0 & 18.6 & 1.0 & 12.0 & 44.0 & 98.0 & 860.0 & 73.3 \\
\midrule
\multirow{2}{*}{\centering Debates}
& SIM & 300 & 4.0 & 4.0 & 4.0 & 4.0 & 4.0 & 4.0 & 1.0 & 1.0 & 6.0 & 22.0 & 133.0 & 16.4 \\
& IQ2 & 350 & 8.0 & 9.0 & 10.0 & 12.0 & 17.0 & 10.7 & 1.0 & 4.0 & 13.0 & 51.0 & 1394.0 & 47.3 \\
\bottomrule
\end{tabular}
\caption{Training Set} \label{tab:stat-train-detail}
\end{subtable}
\caption{Detailed statistics of \cocoevalbench.} \label{tab:statistics-detail}
\end{table*}

\subsection{Benchmark Creation}

Before the process described in Section~\ref{sec:cocoevalbench}, we clean up undesirable artifacts in QMSum Product and Academic. QMSum Product and Academic include transcripts that preserve details of spoken meetings and include disfluencies, fillers, unnecessary repetitions, and special tokens. To avoid the artifacts from influencing our analysis, we clean the utterances by instructing o4-mini to remove them while preserving the remaining content.

In most cases, this clean-up process simply removes special tokens and incomplete sentences containing artifacts, so we do not expect the choice of clean-up model to strongly influence downstream experiments, as long as sufficiently strong models are used. To assess the impact of the clean-up model selection, we conducted a comparison using GPT-4.1, Claude Sonnet 4.6, and o4-mini on 300 turns from QMSum Product and Academic. We manually examined 41 turns where the token-level edit distance of any pair exceeded 2 and the cosine similarity between embeddings~(text-embedding-3-large) for any pair was below 0.85. Even in these cases, we find that the cleaned sentences are mostly semantically similar.

\subsection{Preprocessing for SFT}
\begin{figure}[t]
    \centering
    \includegraphics[width=\linewidth]{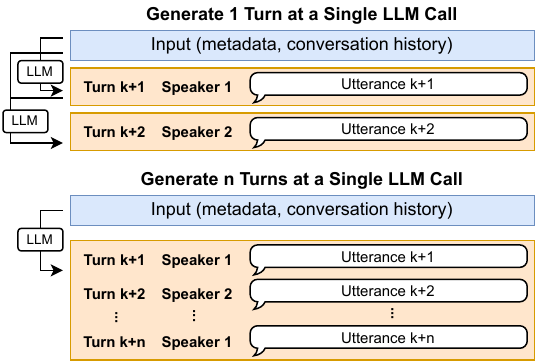}
    \caption{Conversation simulation under one-turn and multi-turn generation per LLM call.}
    \label{fig:generating-multiple-turns}
\end{figure}

For evaluation in Section~\ref{sec:results-finetuning}, we fine-tuned GPT-4.1 on the training set of \cocoevalbench{} with supervised fine-tuning for one epoch with the default parameters in Microsoft Azure. We create SFT datasets for three settings that generate 1, 5, or 30 turns per call, as illustrated in Figure~\ref{fig:generating-multiple-turns}. For the settings that generate 1 or 5 turns per call, we randomly select starting turns for each instance, and the training data size is 1,450 for all settings.

\subsection{Input Length and Token Usage}

In the conversation continuation task in \cocoevalbench, the input includes approximately 4k~tokens, and the output includes about 50 tokens per turn on average. Therefore, for example, when generating 30 turns under the setting of generating one turn per LLM call, the total input tokens amount to approximately $4\text{k} \times 30 = 120\text{k}$~tokens, and the total output tokens amount to about $50 \times 30 = 1.5\text{k}$~tokens. The evaluation set of \cocoevalbench{} includes $50 \times 6 = 300$~conversations, so in total this setting requires approximately $120\text{k} \times 300 \simeq 36\text{M}$ input tokens and $1.5\text{k} \times 300 = 450\text{k}$~output tokens.

\subsection{Unbiased Estimator} \label{appendix:one-continuation}

Each instance in \cocoevalbench{} provides a single realized human continuation~$r$ for a given context~$(M, S, x)$ because it is created from real-world human conversations. However, since human conversation is inherently stochastic, alternative human continuations of the same context could plausibly exhibit different frequencies of inconsistent and uncollaborative behaviors. We therefore do not assume that a single human continuation is representative of the underlying human distribution for its context, and we do not expect a single human continuation and a single LLM-simulated continuation to match in their behavior frequencies. Instead, all comparisons in Section~\ref{sec:results} are made between averages computed over the entire evaluation set, as shown in Figure~\ref{fig:bargraph}.

Statistically, averaging over the dataset with a single reference continuation per context provides an unbiased estimate of the expected behavior frequency. Let $T$ be a behavior type, $D$ a domain, $P_D$ the distribution of contexts~$z = (M, S, x)$ in~$D$, $P(\cdot \mid z)$ the conditional distribution of continuations given a context~$z$, and $f_T(r)$ the function that counts occurrences of behavior type~$T$ in a continuation~$r$. The expected behavior frequency is

{\small
\[
\mathbb{E}_{z \sim P_D}
\bigl[
\mathbb{E}_{r \sim P(\cdot \mid z)} [f_T(r)]
\bigr].
\]
}

\noindent
With $N$ sampled contexts and $K$ continuations~$r_{i,1}, \ldots, r_{i,K}$ per context~$z_i$, an unbiased estimator of this quantity is

{\small
\[
\frac{1}{N} \sum_{i=1}^{N}
\left(
\frac{1}{K} \sum_{j=1}^{K} f_T(r_{i,j})
\right).
\]
}

\noindent
With a single reference continuation per context, the corresponding estimator is

{\small
\[
\frac{1}{N} \sum_{i=1}^{N} f_T(r_{i,1}).
\]
}

\noindent
Averaging this single-reference estimate over the sampled contexts therefore remains unbiased for the expected behavior frequency.

\section{LLM-as-a-Judge} \label{appendix:llm-as-a-judge-agreement}

This section provides additional details for Section~\ref{sec:llm-as-a-judge-agreement} on the agreement between human evaluation and LLM-as-a-Judge in \cocoeval{}.

\subsection{Setup of Human Evaluation} \label{appendix:llm-as-a-judge-agreement-setting}

Five PhD students working on NLP research conducted the human evaluation. They were provided with the same instructions as those used in the prompt for the LLM-as-a-Judge, and the annotators were not informed whether the conversations were human conversations or LLM-simulated ones. The inputs~$(M, S, x)$ are randomly selected from the evaluation set of \cocoevalbench{} and evenly sampled from instances created from the six source datasets.

\paragraph{Overall evaluation.}
The human annotators manually evaluate the overall consistency and collaborativeness of 54 conversation continuations generated by humans, GPT-4.1, and GPT-4.1 with SFT on human conversations, where LLMs generate 30 turns per call. They assign scores on a 1--10 scale by referring to the instructions and rubrics provided to LLM-as-a-Judge.

\paragraph{Fine-grained evaluation.}
The human annotators manually detect the 10~types of inconsistent and uncollaborative behaviors in 1,080 turns from 36~human conversations and conversation continuations simulated by GPT-4.1~(generating 30 turns per call). The annotators provide the indices of turns that exhibit each type of behavior.

\paragraph{Inter-human agreement.}
We measure inter-human agreement by the average leave-one-out Spearman's rank correlation. For each of the five annotators, we correlate that annotator's scores with the mean of the remaining four annotators, and we average the resulting five correlations. We assess statistical significance with a one-sided bootstrap over conversations~(1,000 resamples), treating a correlation as significant when the lower 5th percentile of the bootstrap distribution lies above zero. The test is one-sided because only a positive correlation indicates agreement.

\begin{table}[t]
\centering
\small
\begin{tabular}{crr}
\toprule
Model & Consistency & Collaborativeness \\
\midrule
o4-mini & 0.546* & 0.586* \\
Claude Opus 4 & 0.223\phantom{*} & 0.177\phantom{*} \\
GPT-5.1~(Instant) & 0.284* & 0.306* \\
GPT-5.1~(High) & 0.240* & 0.437* \\
\bottomrule
\end{tabular}
    \caption{Average leave-one-out Spearman's rank correlation between LLM-as-a-Judge and human evaluation in the overall evaluation in \cocoeval~(Appendix~\ref{appendix:llm-as-a-judge-agreement-setting}). $^*$: $p <$ 0.05~(one-sided bootstrap).}
    \label{tab:llm-as-a-judge-model-choice-overall}
\end{table}

\begin{table*}[t]
\centering
\scriptsize
\begin{tabular}{lrrrrrrrrrr}
\toprule
\multirow{2}{*}{Category} & \multicolumn{2}{c}{Human} & \multicolumn{2}{c}{o4-mini} & \multicolumn{2}{c}{GPT-5.1~(Instant)} & \multicolumn{2}{c}{GPT-5.1~(High)} & \multicolumn{2}{c}{Claude Opus 4} \\
\cmidrule(lr){2-3}\cmidrule(lr){4-5}\cmidrule(lr){6-7}\cmidrule(lr){8-9}\cmidrule(lr){10-11}
 & \multicolumn{1}{c}{$\rho$} & \multicolumn{1}{c}{\# turns} & \multicolumn{1}{c}{$\rho$} & \multicolumn{1}{c}{\# turns} & \multicolumn{1}{c}{$\rho$} & \multicolumn{1}{c}{\# turns} & \multicolumn{1}{c}{$\rho$} & \multicolumn{1}{c}{\# turns} & \multicolumn{1}{c}{$\rho$} & \multicolumn{1}{c}{\# turns} \\
\midrule
Logical Contradiction & 0.211\phantom{$^*$} & 6.0 & 0.044\phantom{$^*$} & 4 & 0.286$^*$ & 15 & 0.286$^*$ & 12 & 0.106\phantom{$^*$} & 6 \\
Factual Inaccuracy & 0.256\phantom{$^*$} & 2.4 & 0.253\phantom{$^*$} & 2 & 0.156\phantom{$^*$} & 6 & ---\phantom{$^*$} & 0 & -0.075\phantom{$^*$} & 3 \\
Misunderstanding & 0.001\phantom{$^*$} & 4.6 & 0.208$^*$ & 6 & ---\phantom{$^*$} & 0 & ---\phantom{$^*$} & 0 & 0.276$^*$ & 9 \\
Redundant Information & 0.200$^*$ & 10.4 & 0.185\phantom{$^*$} & 12 & 0.227\phantom{$^*$} & 66 & 0.355$^*$ & 79 & 0.401$^*$ & 48 \\
Repetition & 0.370$^*$ & 18.4 & 0.613$^*$ & 23 & 0.270$^*$ & 91 & 0.354$^*$ & 72 & 0.315$^*$ & 72 \\
\midrule
Persistent Disagreement & 0.358\phantom{$^*$} & 4.4 & 0.315$^*$ & 2 & ---\phantom{$^*$} & 0 & ---\phantom{$^*$} & 0 & 0.594$^*$ & 32 \\
Interruption & 0.728$^*$ & 33.4 & 0.673$^*$ & 12 & 0.363$^*$ & 6 & ---\phantom{$^*$} & 0 & 0.786$^*$ & 22 \\
Off Topic & 0.423$^*$ & 11.6 & 0.478$^*$ & 15 & 0.183$^*$ & 1 & 0.363$^*$ & 8 & 0.564$^*$ & 29 \\
Under Answering & 0.356$^*$ & 16.8 & 0.661$^*$ & 38 & 0.580$^*$ & 17 & 0.542$^*$ & 15 & 0.531$^*$ & 43 \\
Unclear Intent & 0.212$^*$ & 9.8 & 0.672$^*$ & 42 & 0.246$^*$ & 6 & 0.367$^*$ & 18 & 0.553$^*$ & 23 \\
\bottomrule
\end{tabular}
\caption{Agreement between human evaluation and LLM-as-a-Judge using different backbone LLMs in \cocoeval's fine-grained evaluation~(the average leave-one-out Spearman's $\rho$). $^*$: $p <$ 0.05~(one-sided bootstrap).}
\label{tab:llm-as-a-judge-model-choice-fine-grained}
\end{table*}

\subsection{Backbone LLMs for LLM-as-a-Judge} \label{appendix:llm-as-a-judge-other-models}

Our main analysis in Section~\ref{sec:results} uses o4-mini as LLM-as-a-Judge, as it shows reasonable alignment with human evaluation. We also evaluate GPT-5.1 Instant and High~\citep{gpt-5-1} and Claude Opus 4~\citep{claude4} as LLM-as-a-Judge models to analyze the influence of backbone LLMs for LLM-as-a-Judge.

First, Table~\ref{tab:llm-as-a-judge-model-choice-overall} shows the Spearman's rank correlation between human evaluation and the LLM-as-a-Judge results. o4-mini shows the strongest correlation with human evaluation, while other models also exhibit positive correlation.

Next, Table~\ref{tab:llm-as-a-judge-model-choice-fine-grained} presents the results in the fine-grained evaluation. Spearman's rank correlation shows similar trends across all models, but the number of detected turns varies substantially. Specifically, while o4-mini mostly shows trends similar to human evaluation, GPT-5.1 underdetects some categories, including Misunderstanding and Persistent Disagreement, and overdetects Redundant Information and Repetition. Claude Opus 4 tends to overdetect most categories.

In summary, o4-mini exhibits reasonable correlation with human evaluation in both evaluation schemes, and we select o4-mini for LLM-as-a-Judge in our main analysis. We consider that prompt engineering and careful calibration against human evaluation are needed to better align other LLM-as-a-Judge models with human evaluation.

\subsection{Coverage of Fine-grained Evaluation}

To verify that the 10 types of behaviors used in the fine-grained evaluation of \cocoeval{} cover most inconsistent and uncollaborative behaviors in human conversations, we instructed the human annotators to identify any behaviors not captured by these categories when conducting human evaluation for the fine-grained evaluation. In the reviewed conversations, the annotators did not identify any behaviors outside the 10~types of behaviors.

\section{Detailed Settings}

We provide detailed experimental settings.

\subsection{Model Access}

\paragraph{Simulation.}
We use \texttt{gpt-4.1 2025-04-14} and \texttt{gpt-5.1-chat 2025-11-13} via Microsoft Azure, and \texttt{claude-opus-4-20250514} via Claude API.

\paragraph{LLM-as-a-Judge and benchmark creation.}
We use \texttt{o4-mini 2025-04-16} for LLM-as-a-Judge via Microsoft Azure.

\begin{figure}[t!]
    \centering
\begin{subfigure}{\linewidth}
    \includegraphics[width=\linewidth,trim={.3cm 3.8cm 0 0},clip]{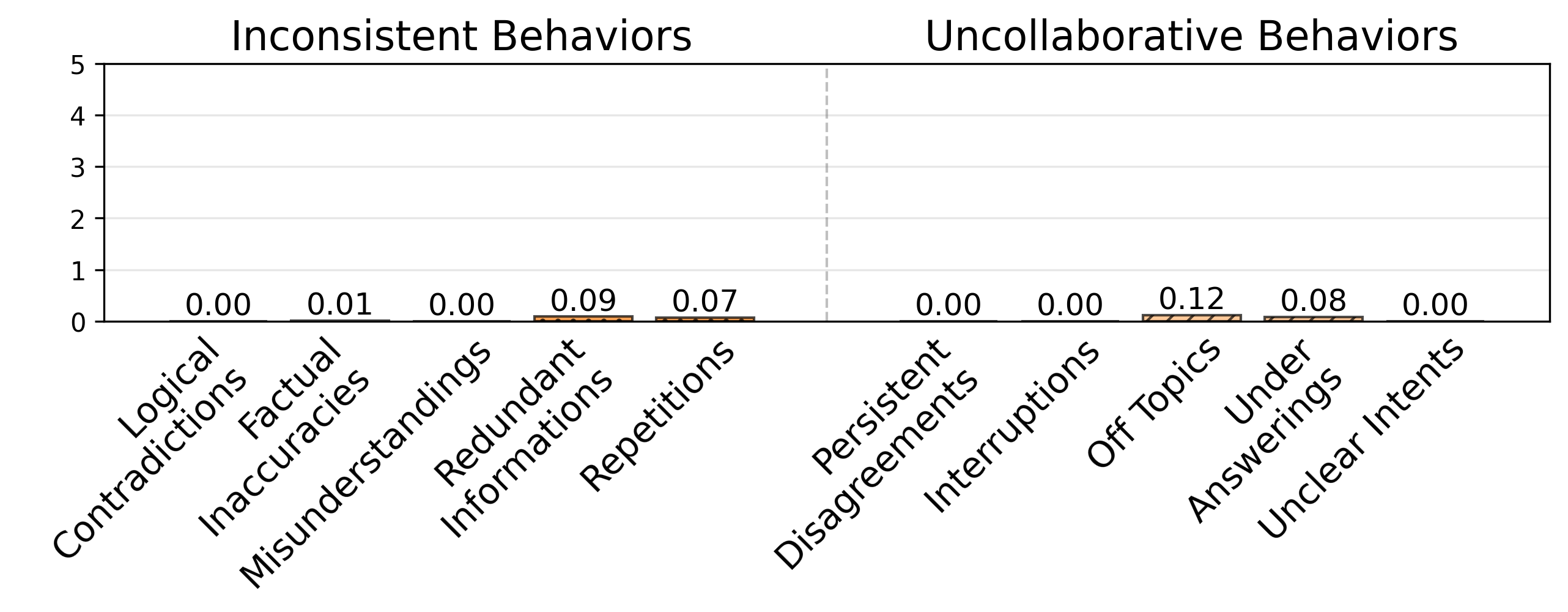}
    \caption{Vanilla prompting with 5 turns generated per LLM call}
\end{subfigure}
\vskip .35em
\begin{subfigure}{\linewidth}
    \includegraphics[width=\linewidth,trim={.3cm 3.8cm 0 .75cm},clip]{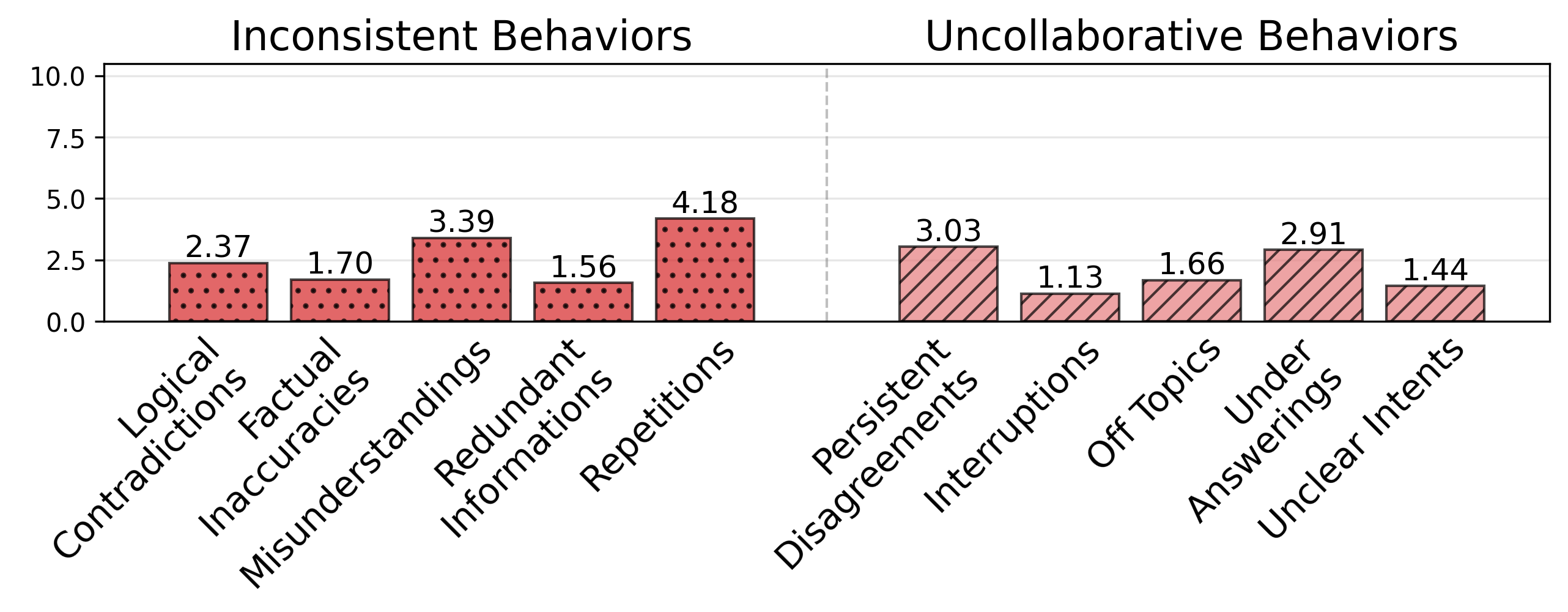}
    \caption{Taxonomy-guided prompting with 5 turns per call}
\end{subfigure}
\vskip .35em
\begin{subfigure}{\linewidth}
    \includegraphics[width=\linewidth,trim={.3cm .3cm 0 .75cm},clip]{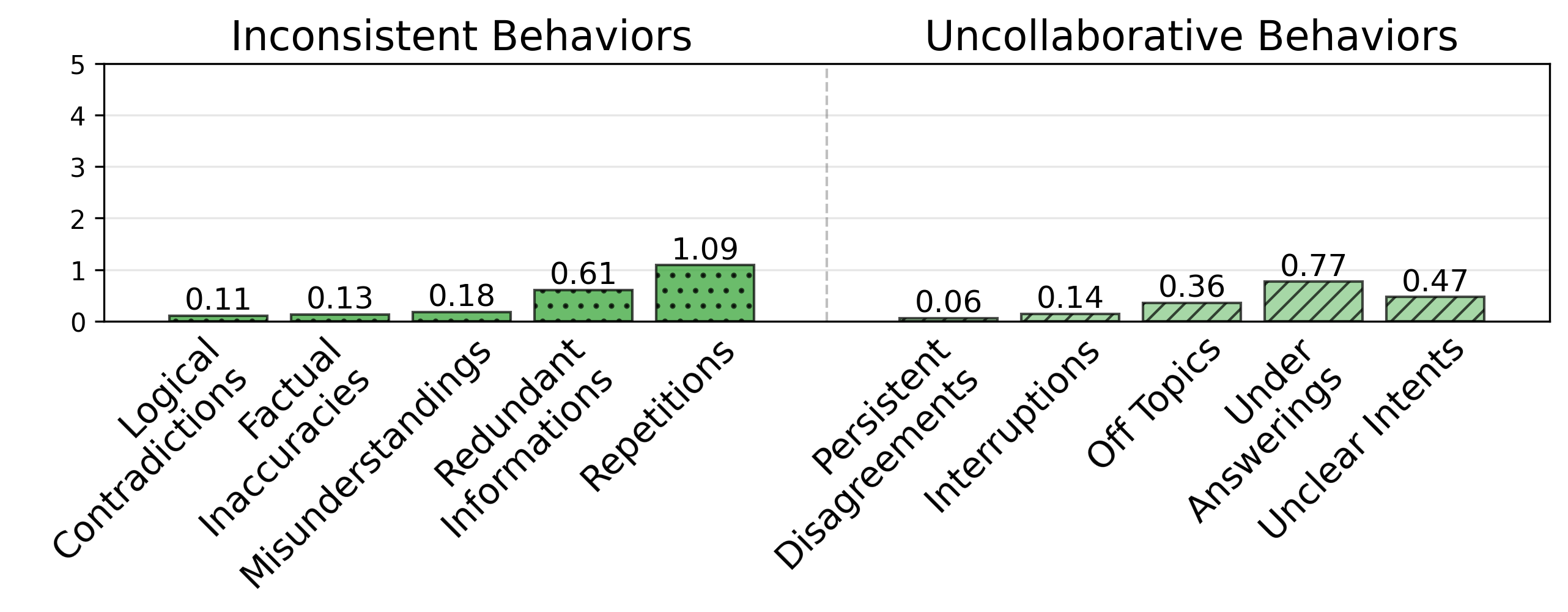}
    \caption{SFT with 5 turns generated per LLM call}
\end{subfigure}
    \caption{\cocoeval's fine-grained evaluation results for conversations simulated by GPT-4.1 with 5 turns generated per call, evaluated by o4-mini.}
    \label{fig:bargraph-gpt4.1-5turns}
\end{figure}

\begin{figure}[t!]
    \centering
\begin{subfigure}{\linewidth}
    \includegraphics[width=\linewidth,trim={.3cm 3.8cm 0 0},clip]{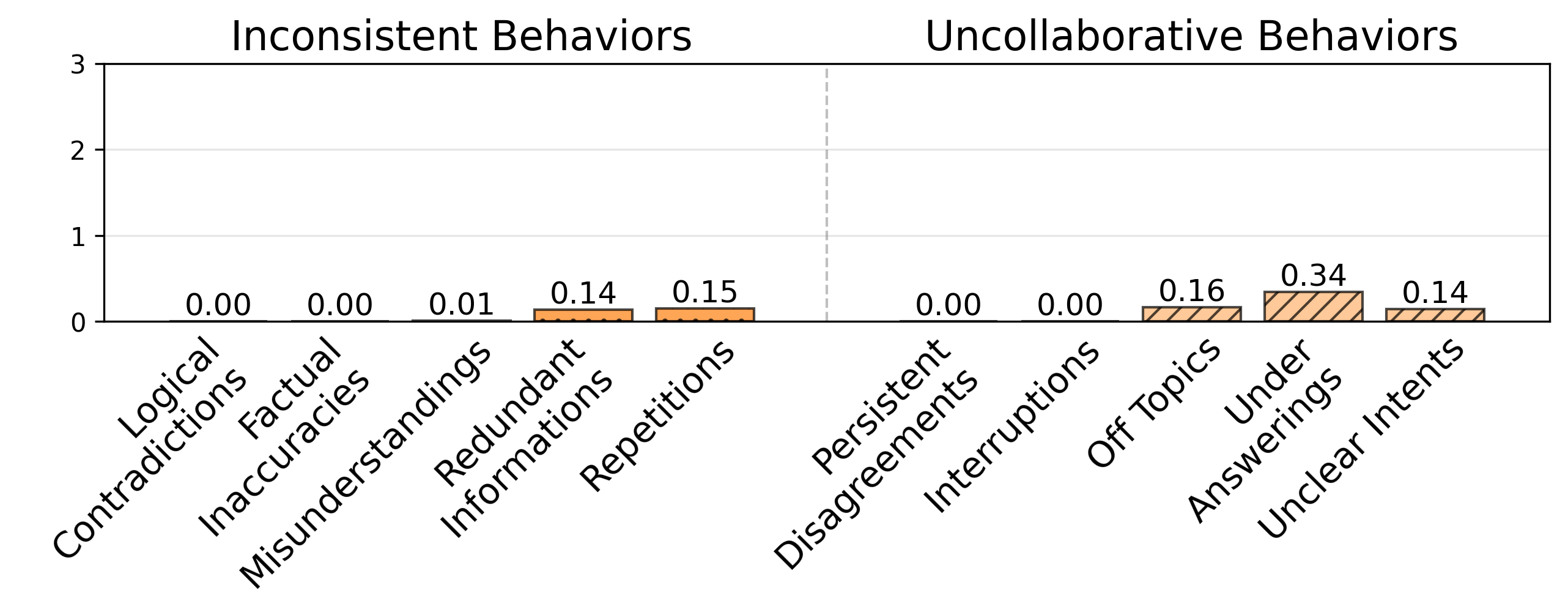}
    \caption{Vanilla prompting with 30 turns generated per LLM call}
\end{subfigure}
\vskip .35em
\begin{subfigure}{\linewidth}
    \includegraphics[width=\linewidth,trim={.3cm 3.8cm 0 .75cm},clip]{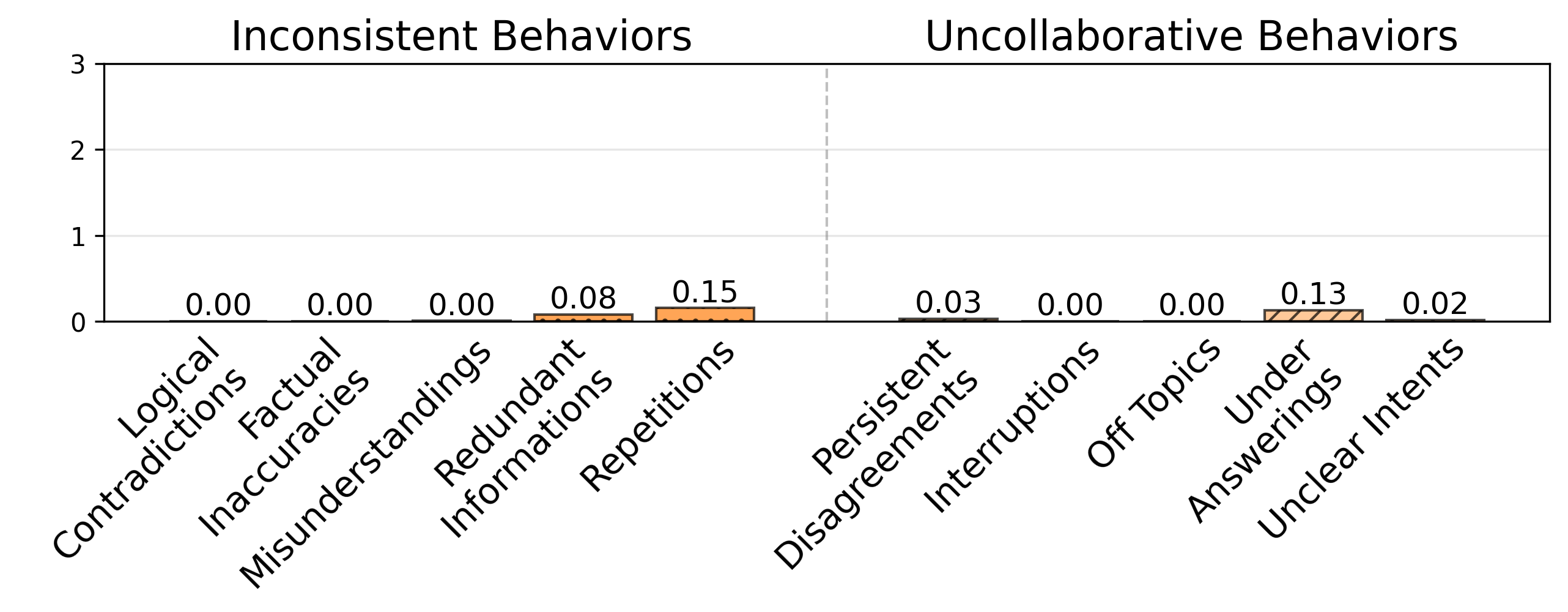}
    \caption{Vanilla prompting with 5 turns generated per LLM call} \label{figure:gpt-5.1-vanilla-5turns}
\end{subfigure}
\vskip .35em
\begin{subfigure}{\linewidth}
    \includegraphics[width=\linewidth,trim={.3cm 3.8cm 0 .75cm},clip]{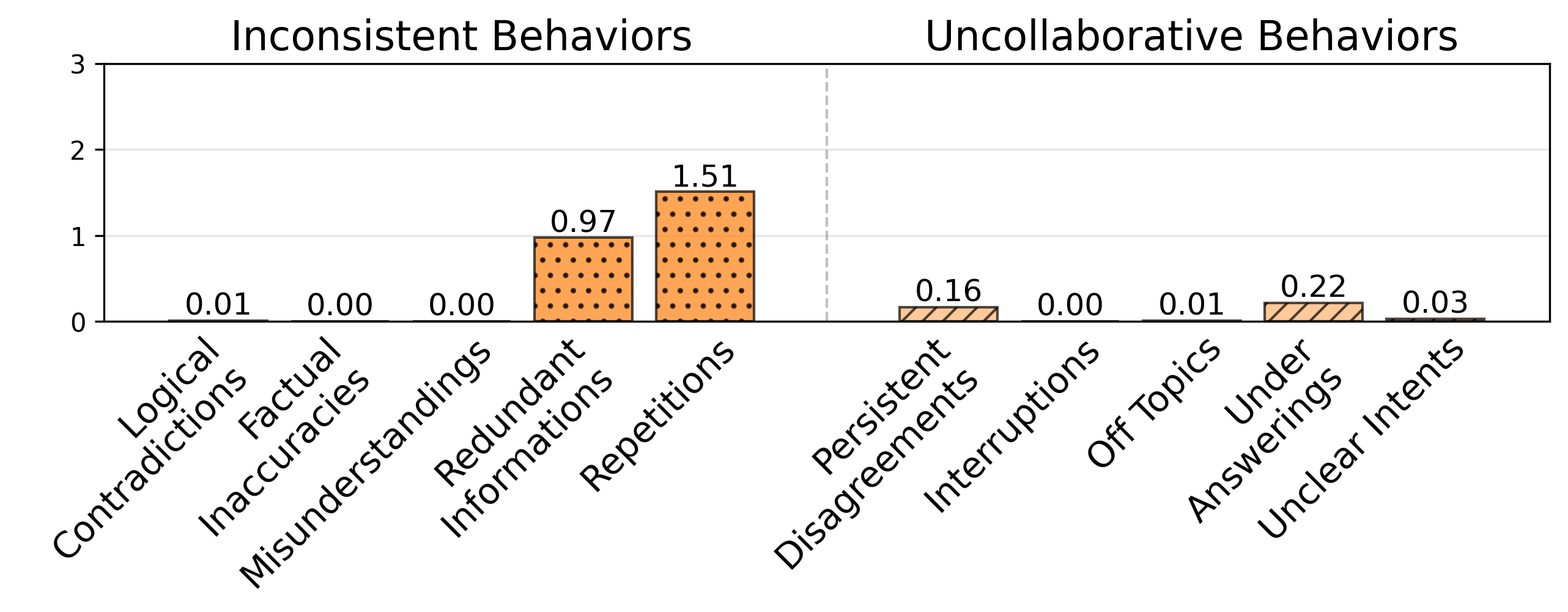}
    \caption{Vanilla prompting with 1 turn generated per LLM call}
    \label{figure:gpt-5.1-vanilla-1turn}
\end{subfigure}
\vskip .35em
\begin{subfigure}{\linewidth}
    \includegraphics[width=\linewidth,trim={.3cm 3.8cm 0 .75cm},clip]{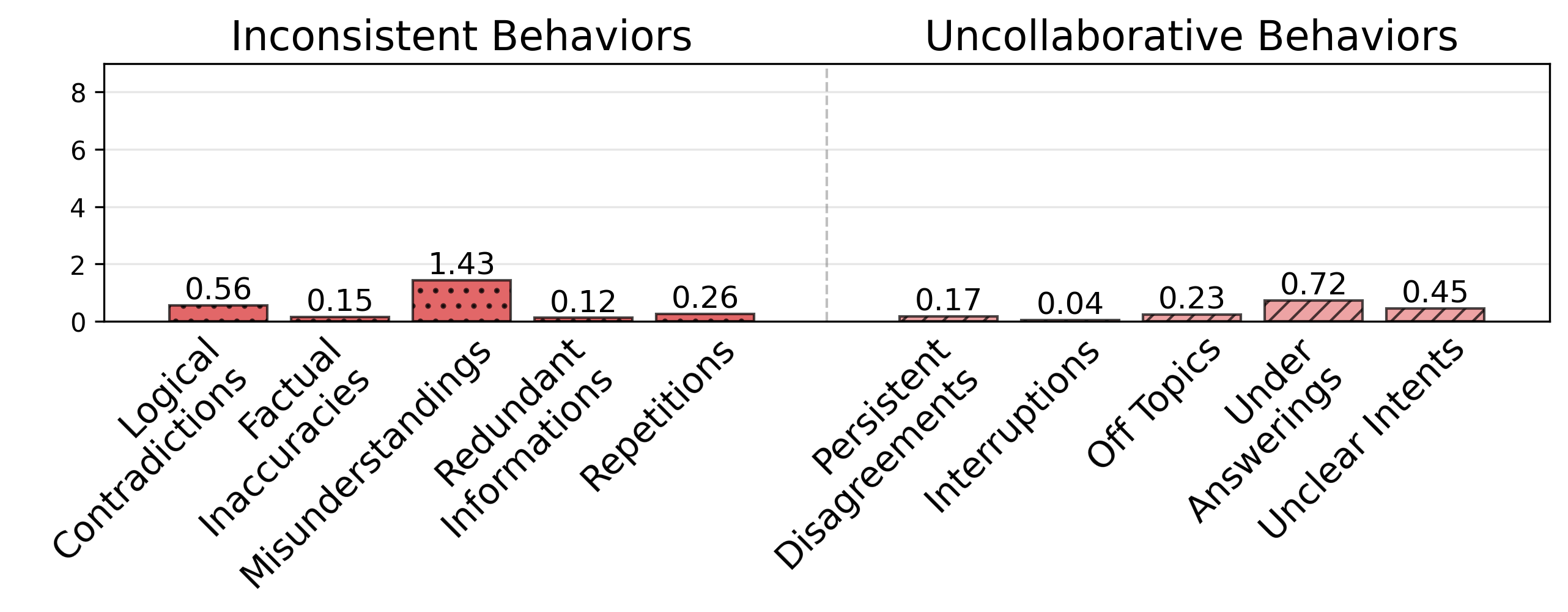}
    \caption{Taxonomy-guided prompting with 30 turns per call}
\end{subfigure}
\vskip .35em
\begin{subfigure}{\linewidth}
    \includegraphics[width=\linewidth,trim={.3cm 3.8cm 0 .75cm},clip]{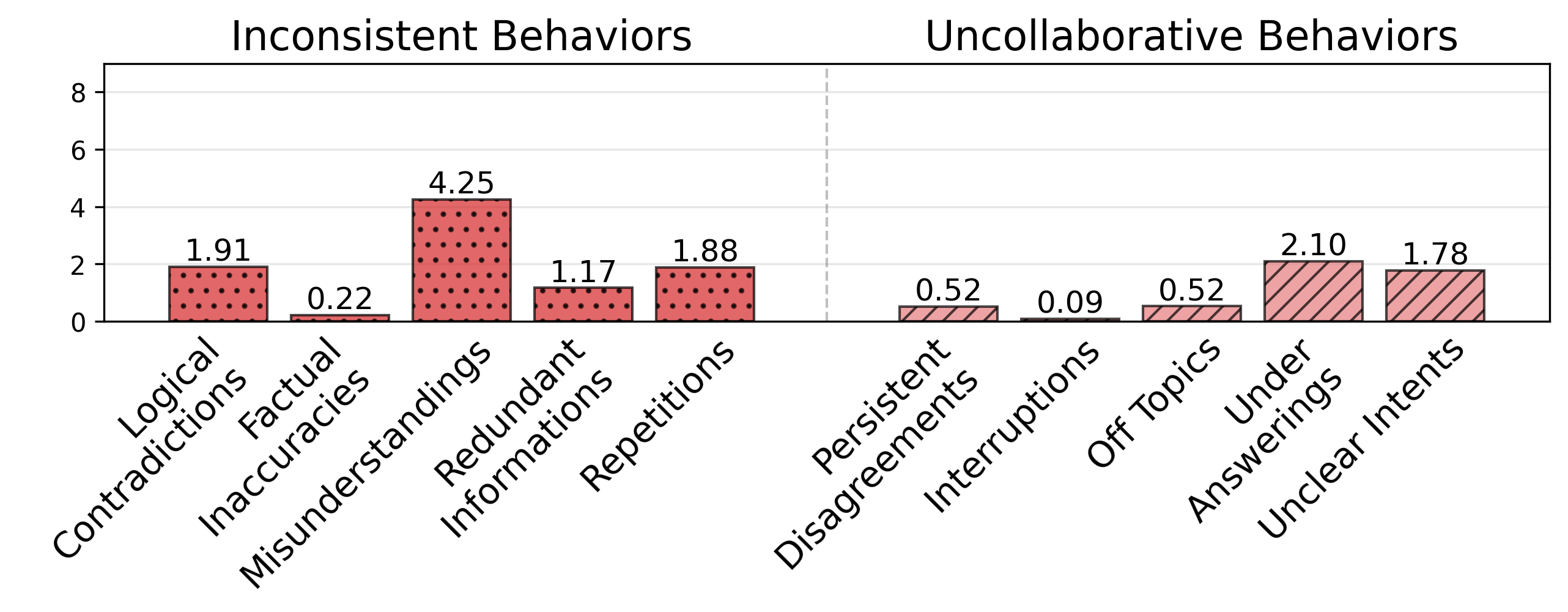}
    \caption{Taxonomy-guided prompting with 5 turns per call}
\end{subfigure}
\vskip .35em
\begin{subfigure}{\linewidth}
    \includegraphics[width=\linewidth,trim={.3cm .3cm 0 .75cm},clip]{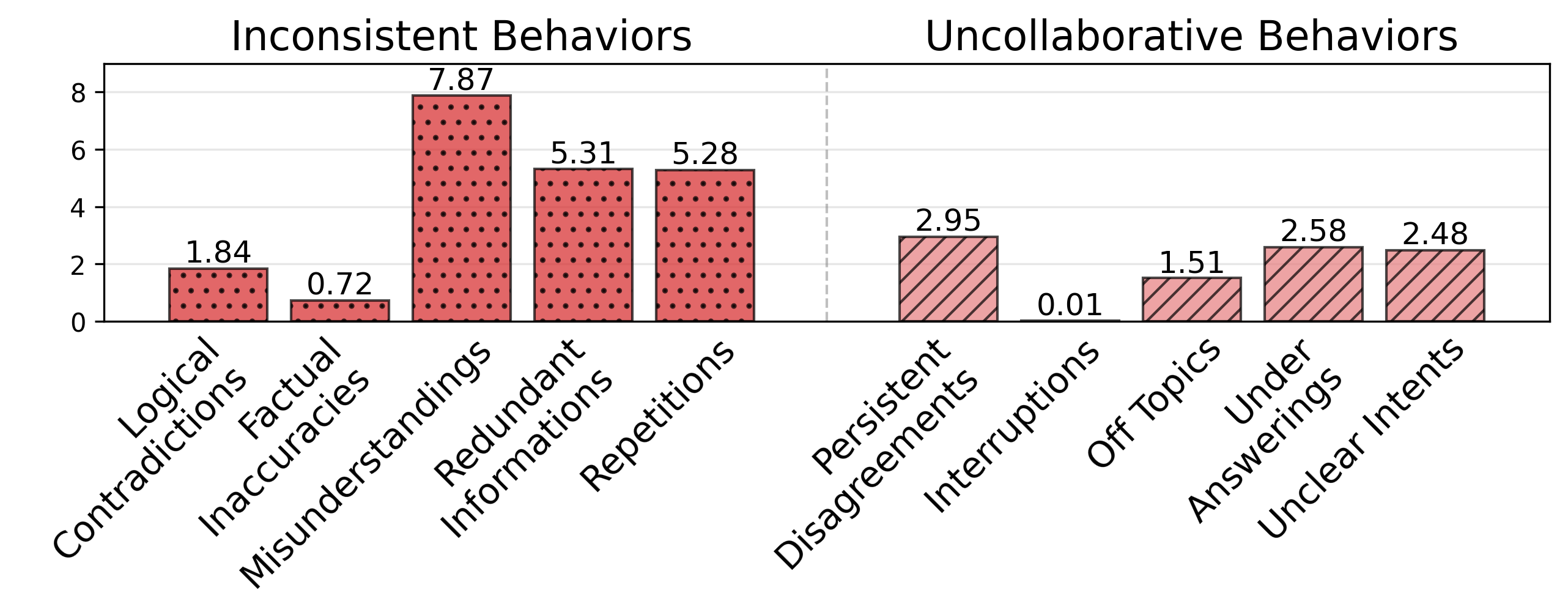}
    \caption{Taxonomy-guided prompting with 1 turn per call} \label{fig:gpt-5.1-taxonomy-guided-1-turn}
\end{subfigure}
    \caption{\cocoeval's fine-grained evaluation results for GPT-5.1, evaluated by o4-mini.}
    \label{fig:bargraph-gpt5.1}
\end{figure}

\begin{figure}[t!]
    \centering
\begin{subfigure}{\linewidth}
    \includegraphics[width=\linewidth,trim={.3cm 3.8cm 0 0},clip]{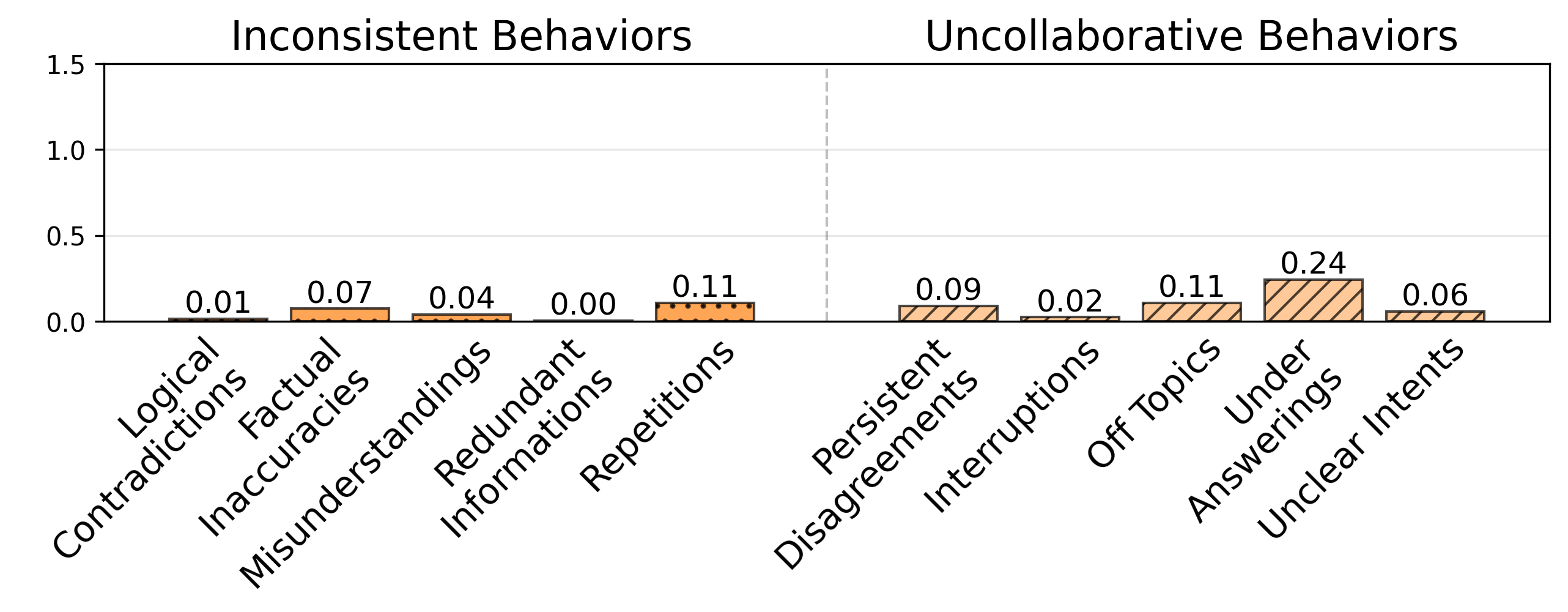}
    \caption{Vanilla prompting with 5 turns generated per LLM call}
\end{subfigure}
\vskip .35em
\begin{subfigure}{\linewidth}
    \includegraphics[width=\linewidth,trim={.3cm 3.8cm 0 .75cm},clip]{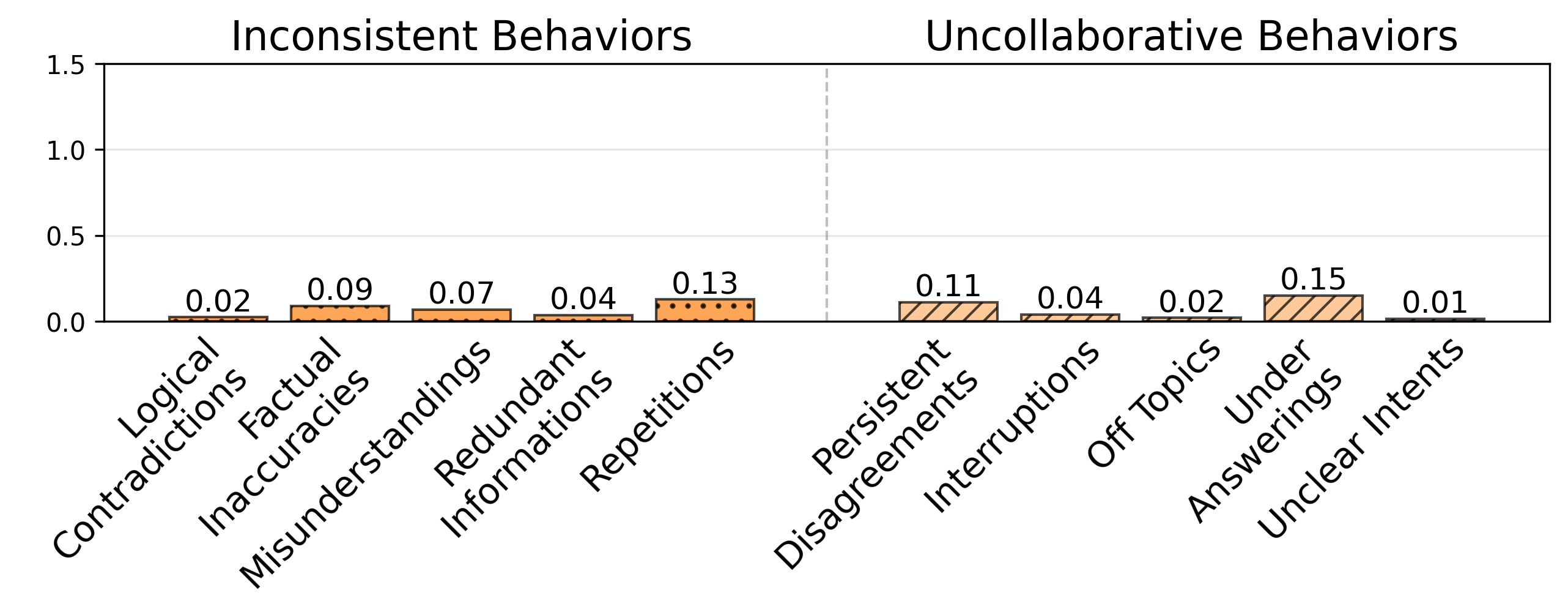}
    \caption{Vanilla prompting with 1 turn generated per LLM call} \label{fig:claude-vanilla-1turn}
\end{subfigure}
\vskip .35em
\begin{subfigure}{\linewidth}
    \includegraphics[width=\linewidth,trim={.3cm 3.8cm 0 .75cm},clip]{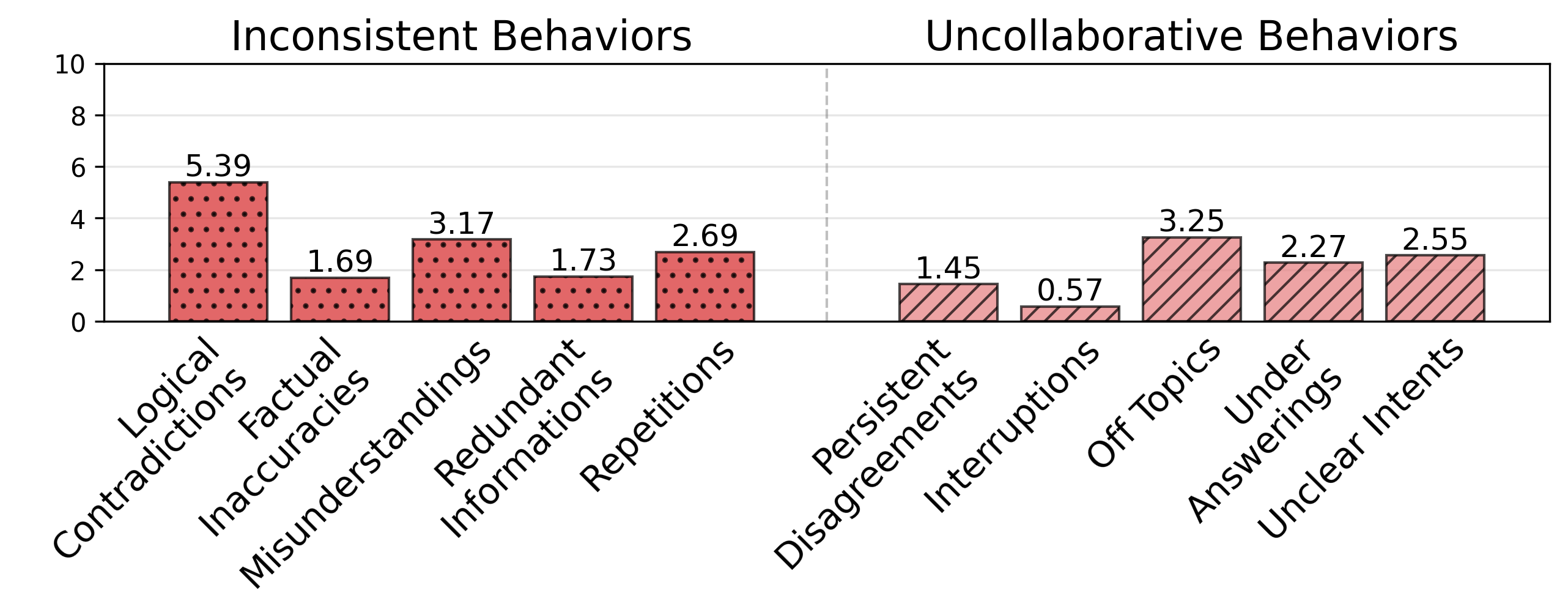}
    \caption{Taxonomy-guided prompting with 5 turns per call}
\end{subfigure}
\vskip .35em
\begin{subfigure}{\linewidth}
    \includegraphics[width=\linewidth,trim={.3cm .3cm 0 .75cm},clip]{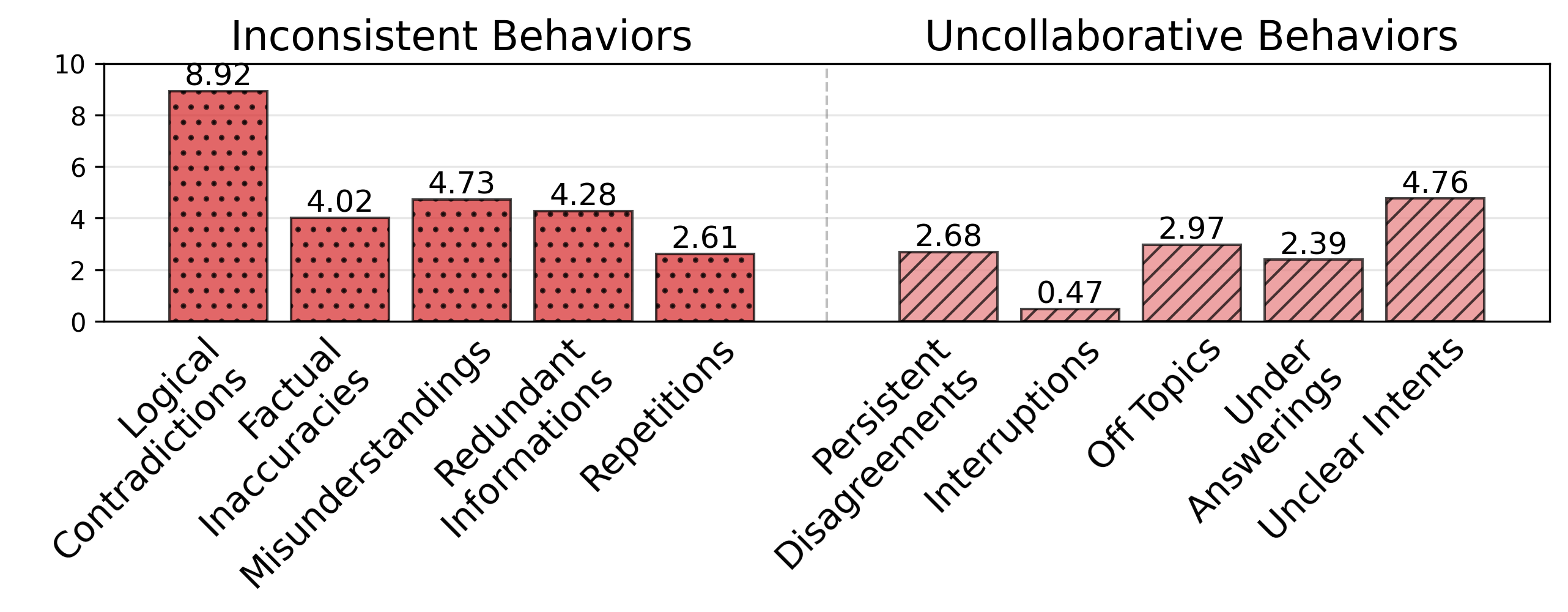}
    \caption{Taxonomy-guided prompting with 1 turn per call}
\end{subfigure}
    \caption{\cocoeval's fine-grained evaluation results for Claude~Opus~4, evaluated by o4-mini.}
    \label{fig:bargraph-claude}
\end{figure}

\begin{table*}[t]
\centering
\tiny
\setlength{\tabcolsep}{4pt}
\begin{subtable}{\linewidth}
\centering
\begin{tabular}{lrrrrrrrrrr}
\toprule
Dataset & \shortstack{Logical\\Contradiction} & \shortstack{Factual\\Inaccuracy} & \shortstack{Misunder-\\standing} & \shortstack{Redundant\\Information} & \shortstack{Repetition} & \shortstack{Persistent\\Disagreement} & \shortstack{Interruption} & \shortstack{Off Topic} & \shortstack{Under\\Answering} & \shortstack{Unclear Intent} \\
\midrule
QMSum~Academic & 0.10 & 0.04 & 0.16 & 0.29 & 0.57 & 0.00 & 0.00 & 0.98 & 0.43 & 0.63 \\
QMSum~Product & 0.64 & 0.16 & 0.50 & 0.38 & 1.28 & 0.14 & 0.00 & 0.62 & 1.26 & 1.70 \\
QMSum~Committee & 0.06 & 0.04 & 0.16 & 0.12 & 0.18 & 0.04 & 0.22 & 0.04 & 2.56 & 0.38 \\
NCPC & 0.02 & 0.02 & 0.14 & 0.12 & 0.20 & 0.00 & 0.12 & 0.26 & 0.04 & 0.40 \\
SIM & 0.06 & 0.38 & 0.30 & 1.20 & 1.46 & 0.04 & 0.20 & 2.10 & 2.12 & 4.22 \\
IQ2 & 0.12 & 0.22 & 0.36 & 0.20 & 0.54 & 0.34 & 1.80 & 0.82 & 1.02 & 0.74 \\
\midrule
Average & 0.17 & 0.14 & 0.27 & 0.38 & 0.71 & 0.09 & 0.39 & 0.80 & 1.24 & 1.35 \\
\bottomrule
\end{tabular}
\caption{Human Conversations}
\end{subtable}
\begin{subtable}{\linewidth}
\centering
\begin{tabular}{lrrrrrrrrrr}
\toprule
Dataset & \shortstack{Logical\\Contradiction} & \shortstack{Factual\\Inaccuracy} & \shortstack{Misunder-\\standing} & \shortstack{Redundant\\Information} & \shortstack{Repetition} & \shortstack{Persistent\\Disagreement} & \shortstack{Interruption} & \shortstack{Off Topic} & \shortstack{Under\\Answering} & \shortstack{Unclear Intent} \\
\midrule
QMSum~Academic & 0.00 & 0.00 & 0.00 & 1.38 & 0.50 & 0.00 & 0.00 & 0.80 & 0.00 & 0.00 \\
QMSum~Product & 0.00 & 0.00 & 0.00 & 0.84 & 0.60 & 0.00 & 0.00 & 0.60 & 0.00 & 0.00 \\
QMSum~Committee & 0.00 & 0.00 & 0.00 & 0.02 & 0.10 & 0.00 & 0.00 & 0.00 & 0.00 & 0.00 \\
NCPC & 0.00 & 0.00 & 0.00 & 1.18 & 0.40 & 0.00 & 0.00 & 0.06 & 0.00 & 0.00 \\
SIM & 0.00 & 0.00 & 0.00 & 3.14 & 2.74 & 0.00 & 0.00 & 0.00 & 0.00 & 0.00 \\
IQ2 & 0.00 & 0.02 & 0.00 & 2.20 & 2.06 & 1.40 & 0.00 & 0.20 & 0.00 & 0.00 \\
\midrule
Average & 0.00 & 0.00 & 0.00 & 1.46 & 1.07 & 0.23 & 0.00 & 0.28 & 0.00 & 0.00 \\
\bottomrule
\end{tabular}
\caption{GPT-4.1 (generating 1 turn per call)}
\end{subtable}
\begin{subtable}{\linewidth}
\centering
\begin{tabular}{lrrrrrrrrrr}
\toprule
Dataset & \shortstack{Logical\\Contradiction} & \shortstack{Factual\\Inaccuracy} & \shortstack{Misunder-\\standing} & \shortstack{Redundant\\Information} & \shortstack{Repetition} & \shortstack{Persistent\\Disagreement} & \shortstack{Interruption} & \shortstack{Off Topic} & \shortstack{Under\\Answering} & \shortstack{Unclear Intent} \\
\midrule
QMSum~Academic & 0.02 & 0.00 & 0.20 & 6.26 & 9.00 & 0.00 & 0.00 & 0.02 & 0.24 & 0.14 \\
QMSum~Product & 0.14 & 0.02 & 0.22 & 3.30 & 3.90 & 0.02 & 0.00 & 0.14 & 0.46 & 0.68 \\
QMSum~Committee & 0.00 & 0.06 & 0.18 & 1.90 & 2.70 & 0.08 & 0.04 & 0.08 & 2.08 & 0.24 \\
NCPC & 0.06 & 0.02 & 0.04 & 0.86 & 1.60 & 0.00 & 0.04 & 0.62 & 0.02 & 0.00 \\
SIM & 0.10 & 0.22 & 0.14 & 1.02 & 1.80 & 0.16 & 0.04 & 0.64 & 1.72 & 1.66 \\
IQ2 & 0.16 & 0.58 & 0.20 & 1.28 & 3.54 & 0.22 & 0.98 & 0.28 & 0.90 & 0.42 \\
\midrule
Average & 0.08 & 0.15 & 0.16 & 2.44 & 3.76 & 0.08 & 0.18 & 0.30 & 0.90 & 0.52 \\
\bottomrule
\end{tabular}
\caption{GPT-4.1 with SFT on Human Conversations (generating 1 turn per call)}
\end{subtable}
\caption{Fine-grained evaluation results separately shown for six source datasets in \cocoevalbench. Figure~\ref{fig:bargraph} shows bar graphs of the average values.}
\label{tab:fine-grained-results-per-data}
\end{table*}

\begin{table*}[t]
\centering
\tiny
\setlength{\tabcolsep}{3pt}
\begin{subtable}[t]{\textwidth}
\centering
\begin{tabular}{lcccccccccc}
\toprule
Model & \shortstack{Logical\\Contradiction} & \shortstack{Factual\\Inaccuracy} & \shortstack{Misunder-\\standing} & \shortstack{Redundant\\Information} & \shortstack{Repetition} & \shortstack{Persistent\\Disagreement} & \shortstack{Interruption} & \shortstack{Off Topic} & \shortstack{Under\\Answering} & \shortstack{Unclear Intent} \\
\midrule
Human & -- & -- & 0.157 & 0.006 & -0.006 & -- & -- & -0.027 & 0.143 & -0.213 \\
GPT-4.1 (1 turn per call) & -- & -- & -- & 0.018 & -0.081 & -- & -- & 0.259 & -- & -- \\
SFT GPT-4.1 (1 turn per call) & -- & -- & 0.180 & -0.179 & -0.211 & -- & -- & -- & -0.259 & -0.009 \\
\bottomrule
\end{tabular}
\caption{QMSum Academic}
\end{subtable}
\vskip 1em
\begin{subtable}[t]{\textwidth}
\centering
\begin{tabular}{lcccccccccc}
\toprule
Model & \shortstack{Logical\\Contradiction} & \shortstack{Factual\\Inaccuracy} & \shortstack{Misunder-\\standing} & \shortstack{Redundant\\Information} & \shortstack{Repetition} & \shortstack{Persistent\\Disagreement} & \shortstack{Interruption} & \shortstack{Off Topic} & \shortstack{Under\\Answering} & \shortstack{Unclear Intent} \\
\midrule
Human & 0.400 & -0.129 & -0.174 & -0.180 & -0.210 & -0.161 & 0.092 & -0.342 & -0.142 & -0.003 \\
GPT-4.1 (1 turn per call) & -- & -- & -- & 0.080 & 0.241 & -0.218 & -- & 0.192 & -- & -- \\
SFT GPT-4.1 (1 turn per call) & -0.048 & -0.073 & -0.127 & -0.021 & 0.018 & -0.176 & -0.222 & -0.111 & 0.006 & -0.051 \\
\bottomrule
\end{tabular}
\caption{IQ2}
\end{subtable}
\caption{Spearman's rank correlation between the number of conversation participants and inconsistent and uncollaborative behaviors. The table includes results only for behaviors detected by LLM-as-a-Judge~(o4-mini) in more than five out of 50 conversations; otherwise, the result is shown as ``--''.
}
\label{tab:influence-of-participants}
\end{table*}

\subsection{Mitigating Repetition and Redundancy} \label{appendix:repetition-penalty}

We observe a strong tendency in LLM-simulated conversations to produce repetitive and redundant utterances, particularly when conversations are generated one turn at a time and when the simulation LLMs are fine-tuned on human conversations through SFT. To mitigate this issue, we apply the presence penalty parameter available in OpenAI models. Specifically, as shown in Table~\ref{prompt:generation-system}, we instruct simulation LLMs to include the five most recent turns in the response before generating the next turn, and set \texttt{presence\_penalty = 2.0}. Because the copied context becomes part of the response, the presence penalty discourages the model from generating tokens that already appear in the recent dialogue history. While this strategy reduces repetitive and redundant utterances to some extent, it does not resolve this issue, as demonstrated in Section~\ref{sec:results}. These results suggest that repetition and redundancy are inherent biases in LLM-based conversation simulation and remain difficult to fully address.

\section{Additional Results}

We provide additional experimental results.

\subsection{Generalization Analysis} \label{appendix:human-ai-interaction}

Our main experiments target the setting in which an LLM simulates all participants of a multi-party conversation. To demonstrate the applicability of \cocoeval, as well as to examine whether our findings extend beyond this setting, we conduct an additional experiment on human-AI interaction, where an LLM simulates the human user in a two-party conversation with a chatbot.

\paragraph{Setup.}
We use ShareChat~\citep{yan2025sharechat}, a large-scale corpus of authentic user-LLM conversations publicly shared from multiple platforms including ChatGPT and Gemini. We select 25~natural conversations with more than 60~turns, excluding cases such as coding tasks and sessions that consist of isolated independent questions. Following the same protocol as \cocoevalbench, we use the 30~preceding turns as the conversation history and treat the following 30~turns as the reference continuation. We simulate the user turns using GPT-4.1 with the same vanilla prompting setup as in Section~\ref{sec:results}, where one user turn is generated per LLM call because the simulated user and the chatbot alternate, and the chatbot turns are generated with the system prompt ``You are a helpful assistant.'' We apply the fine-grained evaluation only to the user turns, since we are interested in how faithfully the LLM reproduces human conversational behaviors.

\paragraph{Results.}
Table~\ref{tab:human-ai-interaction} shows the results. The trends are consistent with those observed in our main multi-party experiments. Real user turns exhibit all 10~types of inconsistent and uncollaborative behaviors at moderate frequencies. In contrast, LLM-simulated user turns overproduce Redundant Information and Repetition, while almost never producing any of the other eight behavior types.

\begin{table*}[t]
\centering
\tiny
\setlength{\tabcolsep}{4pt}
\begin{tabular}{lrrrrrrrrrr}
\toprule
Continuation & \shortstack{Logical\\Contradiction} & \shortstack{Factual\\Inaccuracy} & \shortstack{Misunder-\\standing} & \shortstack{Redundant\\Information} & \shortstack{Repetition} & \shortstack{Persistent\\Disagreement} & \shortstack{Interruption} & \shortstack{Off Topic} & \shortstack{Under\\Answering} & \shortstack{Unclear Intent} \\
\midrule
Human~(reference) & 0.52 & 0.40 & 0.48 & 0.12 & 0.60 & 0.92 & 0.04 & 0.16 & 0.28 & 0.60 \\
GPT-4.1~(simulated) & 0.00 & 0.00 & 0.00 & 0.76 & 1.36 & 0.20 & 0.00 & 0.00 & 0.00 & 0.00 \\
\bottomrule
\end{tabular}
\caption{Fine-grained evaluation in the human-AI interaction setting, computed over the user turns of 25~conversations from ShareChat~\citep{yan2025sharechat}. Each value is the average number of occurrences per continuation.}
\label{tab:human-ai-interaction}
\end{table*}

\subsection{Additional Fine-grained Evaluation}

This section provides additional results of the fine-grained evaluation in Section~\ref{sec:results}.

\paragraph{Additional bar graphs.}
Figure~\ref{fig:bargraph-gpt4.1-5turns} shows the fine-grained evaluation results of GPT-4.1 when generating 5~turns per LLM call. The frequency of each behavior often falls between those observed in the setups generating 1 and 30~turns per call in Figure~\ref{fig:bargraph}. In addition, Figures~\ref{fig:bargraph-gpt5.1} and \ref{fig:bargraph-claude} show the results of conversations simulated by GPT-5.1 and Claude~Opus~4. The results for these two LLMs mostly align with those for GPT-4.1, but we observe that Claude~Opus~4 with vanilla prompting generates conversations with very few inconsistent and uncollaborative behaviors, even when generating one turn per call~(Figure~\ref{fig:claude-vanilla-1turn}), while GPT-4.1 and GPT-5.1 tend to generate a relatively large number of repetitive and redundant utterances when generating one turn per call~(Figures~\ref{fig:vanilla-1turn}, \ref{figure:gpt-5.1-vanilla-1turn}).

\paragraph{Separate results for six source datasets.}
Table~\ref{tab:fine-grained-results-per-data} shows the fine-grained evaluation results for six source datasets in \cocoevalbench{} separately. These results show two additional findings beyond the results averaged over all datasets, shown in Figure~\ref{fig:bargraph}. First, human conversations exhibit inconsistent and uncollaborative behaviors with varying frequencies across datasets and domains. Second, SFT enables simulated conversations to capture trends in uncollaborative behaviors, such as persistent disagreement and off-topic responses, although it exaggerates redundant information and repetition.

\subsection{Examples of Simulated Continuations}

Tables~\ref{tab:example-reference} and~\ref{tab:example-gpt41} present example continuations for an instance in \cocoevalbench{} created from QMSum Product generated by humans~(reference continuation) and GPT-4.1~(generating 30 turns per call), annotated with fine-grained evaluations by o4-mini. This example shows that the reference continuation contains multiple inconsistent and uncollaborative behaviors, whereas the GPT-4.1 continuation contains only one inconsistent behavior. In addition, the LLM-simulated conversation is more efficient, covering multiple topics within 30 turns and wrapping up the meeting. This example illustrates how LLM simulations fail to reproduce inconsistent and uncollaborative behaviors observed in human conversations and fail to capture the complexity of human social interaction, which often requires many turns to progress.

\subsection{Influence of Number of Participants}

To evaluate the influence of the number of participants in conversations, Table~\ref{tab:influence-of-participants} shows Spearman's correlations between the number of each inconsistent or uncollaborative behavior and the number of participants in the two datasets with relatively high variation in participant counts. We observe that, in most cases, inconsistent and uncollaborative behaviors do not exhibit strong correlations with the number of participants. However, in human conversations from IQ2, Logical Contradiction~(0.400) and Off Topic~(-0.342) show relatively strong correlations. LLM-simulated conversations fail to reproduce this trend, providing further evidence of the mismatch between human and LLM-simulated conversations.

\section{Prompts}

This section presents the prompts used in our experiments and the benchmark creation.

\subsection{\cocoevalbench{} Creation}

We use o4-mini to generate conversational metadata $M$ from the raw utterances in the conversational datasets. Table~\ref{prompt:metadata} shows a part of the metadata extraction prompt. 

\subsection{Conversation Continuation Prompts} \label{appendix:generation-prompt}

Tables~\ref{prompt:generation-system} and \ref{prompt:generation-user} show the vanilla prompt used to simulate conversation continuations using LLMs. Table~\ref{prompt:taxonomy-guided} shows the taxonomy-guided prompt, which includes the list of 10~types of inconsistent and uncollaborative behaviors.

\subsection{Evaluation Prompts} \label{appendix:evaluation-prompt}

Tables~\ref{prompt:overall-evaluation} and \ref{prompt:fine-grained-evaluation} show prompts for LLM-as-a-Judge for the overall and fine-grained evaluation.

\section{Ethical Considerations}

\paragraph{Human annotation.}
Before conducting human evaluation, we manually verified that the provided conversations contain no offensive content. The annotators were informed about this study and provided consent for their annotations to be used.

\paragraph{Dataset.}
Our benchmark is created from public conversational datasets. We do not expect that our benchmark includes information that uniquely identifies individual people, except for politicians included in governmental meetings from QMSum~Committee and NCPC.

\paragraph{The use of LLMs.}
LLMs were used to assist with polishing the writing of this paper and with writing the analysis code.

\begin{table*}[t]
    \centering
    \scriptsize
\begin{tabular}{cM{.15\linewidth}L{.48\linewidth}M{.1\linewidth}M{.1\linewidth}}
\toprule
Turn & Speaker & Content & Inconsistent Behavior & Uncollaborative Behavior \\
\midrule
368 & Industrial Designer & Yes, they can. But they really need the regular chip, you cannot use the simple one. \\
\midrule
369 & User Interface & Okay, but we also already need the regular chip for the sample sensor speaker. & Misunderstanding \\
\midrule
370 & Industrial Designer & Well, I'm not very sure. Maybe that's a different one. & & Under Answering \\
\midrule
371 & User Interface & Oh, okay. \\
\midrule
372 & Project Manager & But do we want the curved design? \\
\midrule
373 & User Interface & Should we review design options? \\
\midrule
374 & Project Manager & Let's discuss the design approach. \\
\midrule
375 & Industrial Designer & I think so. If you stick with the simple, straightforward, not curved design, it's too dull. \\
\midrule
376 & Project Manager & It's too dull. Yeah. & Repetition \\
\midrule
377 & Industrial Designer & I don't think our customers will like it. \\
\midrule
378 & Marketing & Yeah. \\
\midrule
379 & Industrial Designer & And if you take the double-curved design, then you can only use the rubber buttons and you cannot use the scroll-wheels. \\
\midrule
380 & Project Manager & Yeah. Okay. Okay. \\
\midrule
381 & Industrial Designer & So I think this is the best of two worlds. & & Unclear Intent \\
\midrule
382 & Project Manager & Okay, what about the issue I addressed at the beginning of the meeting? The voice function with the infrared issue. \\
\midrule
383 & Industrial Designer & Yes. I can imagine it's very dull to talk to a device if you have to point to another device. \\
\midrule
384 & Project Manager & Yeah. \\
\midrule
385 & Industrial Designer & It's very illogical. \\
\midrule
386 & User Interface & But most infrared remote controls don't have to be pointed at all. & Factual Inaccuracy \\
\midrule
387 & Project Manager & Well. \\
\midrule
388 & Industrial Designer & Well, there has to be some pointing, but not always. \\
\midrule
389 & Project Manager & It also depends on your environment. \\
\midrule
390 & User Interface & Well, if you put your hand in front of it, okay, it won't work, but you can point it just to the other wall. \\
\midrule
391 & Project Manager & Well, it depends on your walls actually. If you have smooth walls, you're probably right. \\
\midrule
392 & User Interface & Yeah. \\
\midrule
393 & Project Manager & But if you have carpets on the wall, which our nature-loving friends probably do have, then it might be a bit more of an issue. \\
\midrule
394 & Industrial Designer & All light gets absorbed, yeah. \\
\midrule
395 & Project Manager & So, we need to consider wall surfaces too. \\
\midrule
396 & Industrial Designer & Yes, because the walls reflect the infrared light. & Logical Contradiction \\
\midrule
397 & Project Manager & Yeah. \\
\bottomrule
\end{tabular}
    \caption{Example reference continuation in an instance from \cocoevalbench{} created from QMSum~Product, annotated with fine-grained evaluation by LLM-as-a-Judge~(o4-mini).}
    \label{tab:example-reference}
\end{table*}

\begin{table*}[t]
    \centering
    \scriptsize
\begin{tabular}{cM{.15\linewidth}L{.48\linewidth}M{.1\linewidth}M{.1\linewidth}}
\toprule
Turn & Speaker & Content & Inconsistent Behavior & Uncollaborative Behavior \\
\midrule
368 & Industrial Designer & Mm-hmm, yes, a regular chip can support scroll-wheel functionality, provided the interface is not too complex. \\
\midrule
369 & Project Manager & So, regular chip for scroll-wheel. Confirmed? \\
\midrule
370 & User Interface & Yeah, that works. No need for advanced chip, right? \\
\midrule
371 & Industrial Designer & Correct. Advanced chip is only necessary if we add voice recognition or speech feedback features. \\
\midrule
372 & Marketing & Younger crowd loves voice commands. But scroll-wheel is cool too. \\
\midrule
373 & Project Manager & Let’s keep voice recognition optional. Main prototype: regular chip, scroll-wheel, push-buttons. \\
\midrule
374 & User Interface & Okay. I’ll mock up button layout with scroll-wheel on the side. \\
\midrule
375 & Industrial Designer & Mm-hmm, please align button zones with ergonomic grip areas. That will improve usability. \\
\midrule
376 & Marketing & Make sure the scroll-wheel is easy to find. Kids like fidgety stuff. \\
\midrule
377 & Project Manager & Roo, deadline for mock-ups is Friday. Sebastian, specs by Monday. \\
\midrule
378 & User Interface & Friday’s fine. I’ll send sketches in PowerPoint. \\
\midrule
379 & Industrial Designer & Okay, I’ll finalize casing dimensions and material specs, including rubber color options. \\
\midrule
380 & Marketing & I’ll get final market feedback on color combos. SpongeBob yellow still trending. \\
\midrule
381 & Project Manager & Ruud, please send market validation by Thursday. I’ll update the shared folder. & Misunderstanding \\
\midrule
382 & User Interface & Do we want the scroll-wheel to control both volume and channel, or just one? \\
\midrule
383 & Industrial Designer & Technically, it can be multi-functional, but simpler is better. I suggest volume only. \\
\midrule
384 & Marketing & Volume’s good. Channel can stay as buttons. Less confusion. \\
\midrule
385 & Project Manager & Agreed. Scroll-wheel for volume. Buttons for channel and power. \\
\midrule
386 & User Interface & Okay, I’ll show that in the mock-up. No LCD, right? \\
\midrule
387 & Industrial Designer & Correct, no LCD. On-screen feedback only, to reduce cost and complexity. \\
\midrule
388 & Marketing & Makes sense. LCD isn’t a must for our target group. \\
\midrule
389 & Project Manager & Let’s confirm: oval rubber body, colorful covers, regular chip, push-buttons, scroll-wheel, optional voice. \\
\midrule
390 & User Interface & Yeah, that’s everything. I’ll get started. \\
\midrule
391 & Industrial Designer & Mm-hmm, I’ll prepare technical drawings and manufacturing notes for review. \\
\midrule
392 & Marketing & I’ll check with the focus group about grip zones and color combos. \\
\midrule
393 & Project Manager & Thanks, team. I’ll summarize decisions and assign tasks in the minutes. \\
\midrule
394 & User Interface & Let me know if you want extra interface ideas. I can add voice command samples. \\
\midrule
395 & Industrial Designer & Okay, if you have time, please include voice command button placement in your mock-up. \\
\midrule
396 & Marketing & I’ll ask the group about favorite voice commands. Maybe ‘Netflix’ button? \\
\midrule
397 & Project Manager & Good idea. Let’s wrap up. Next meeting: prototype review. \\
\bottomrule
\end{tabular}
    \caption{Example continuation generated by GPT-4.1~(generating 30 turns per call) for the instance from \cocoevalbench{} whose reference continuation is shown in Table~\ref{tab:example-reference}, annotated with fine-grained evaluation by LLM-as-a-Judge~(o4-mini).}
    \label{tab:example-gpt41}
\end{table*}

\lstdefinestyle{codebox}{
  basicstyle=\scriptsize\ttfamily,
  frame=single,
  breaklines=true,
  breakatwhitespace=false,
  showstringspaces=false,
  columns=fullflexible,
  keepspaces=true
}

\begin{table*}[t]
\begin{lstlisting}[style=codebox]
```JSON
{
    "task_goal": "Clear description of what needs to be accomplished in the conversation, expressed in a few concise sentences.",
    "task_category": "e.g., project_planning, problem_solving, decision_making",
    "org_context": "Brief description of the organization and situation",
    "participants": [
        {
            "id": "Speaker identifier shown in the metadata",
            "name": "Full name of the participant. If not mentioned, you can provide first name, last name, nickname, alias, or any other relevant identifier. If none of these are available, leave blank.",
            "role": "Job Title, if available. If not available, leave blank.",
            "department": "Department Name, if available. If not available, leave blank.",
            "reporting_to": "id of the participant's manager or supervisor, if available. If not available, leave blank.",
            "expertise_level": "senior, expert, junior, intern, etc.",
            "expertise_areas": [
                "Area of expertise relevant to the conversation (e.g., software development, marketing, finance).", "Another area of expertise."
            ],
            "hidden_information": {
                "communication_style": {
                    "length": "Participant's preferred length of communication (e.g., concise, detailed). Include specific word numbers (e.g., typically 5-10 words).",
                    "style_types": [
                        "formal", "informal", "assertive", "passive", "concise", "elaborative", "technical", "layman"
                    ],
                    "personality_traits": [
                        "A personality trait that may influence the participant's behavior in the conversation (e.g., assertive, passive, analytical, empathetic).", "Another personality trait."
                    ],
                    "details": "A detailed description of the participant's communication style, including tone, vocabulary, sentence structure, and typical phrasing. Note any recurring patterns, preferred ways of expressing agreement or disagreement, use of technical or layman terms, and any unique habits (e.g., frequent use of questions, metaphors, or specific jargon). This should help someone accurately mimic how this participant communicates in conversation."
                },
                "hidden_agenda": "Participant's initial goals or concerns.",
                "knowledge_before_conversation": "Participant's knowledge about the topic before the conversation starts. You should infer this based on the conversation, but do not include any information that is acquired during the conversation.",
                "emotion_before_conversation": "Participant's emotions before the conversation starts (e.g., anxious, excited, neutral).",
                "posture": "Participant's initial posture toward the conversation (e.g., relaxed, tense, open, closed).",
                "beliefs_about_others": {
                    "participant_1": "Participant's beliefs, feelings, and perceptions about Participant 1 before the conversation, including expectations regarding Participant 1's goals, knowledge, and role in the conversation.",
                    "participant_2": "Participant's beliefs, feelings, and perceptions about Participant 2 before the conversation, including expectations regarding Participant 2's goals, knowledge, and role in the conversation."
                }
            }
        },
        {
            ... # Repeat the structure for each participant in the conversation
        }
    ]
}
```
\end{lstlisting}
\caption{Part of the system prompt for generating metadata that specifies the format of the conversational metadata.}
\label{prompt:metadata}
\end{table*}

\begin{table*}[t]
\begin{lstlisting}[style=codebox]
We provide a conversation history involving multiple participants. Your task is to generate {num_generation_turn} responses to simulate the continuation of the conversation.

### Instructions

* Your task is to mimic human-like conversation by referring to the conversation history.
* Your task is to mimic the communication style and content in the metadata and the conversation history.
  * Try to closely mimic the user's style. Your response should match the specified speaker's tone, structure, formatting, and length. Pay attention to stylistic elements such as punctuation, sentence flow, and vocabulary to ensure your reply is consistent with previous messages from that speaker.
  * Conversation metadata may include specific instructions about styles of each participant. You need to follow the instructions to improve the simulation.

### Strict Requirements

* The "speaker" should be selected from the list of participants provided in the metadata. You should not introduce new participants not listed in the metadata.
* Don't finish the conversation. If the current topic concludes, smoothly transition to a related topic to keep the conversation going.

### Output Format

* You first need to copy the last five turns from the conversation history as they are. Then, generate the specified number of new turns in the conversation. Each generated turn should include the turn number, speaker, and the generated response.
* When writing strings in JSON (for example, in "content" fields), ensure that all special characters, especially double quotes ("), are properly escaped using \".

```JSON
{
  "five_previous_turns": [
    {
      "turn_number": first_generated_turn_number - 5,
      "speaker": "speaker_name",
      "content": "Copy the utterance provided in the input, five turns before the first generated turn."
    },
    {
      "turn_number": first_generated_turn_number - 4,
      "speaker": "speaker_name",
      "content": "Copy the utterance provided in the input, four turns before the first generated turn."
    },
    {
      "turn_number": first_generated_turn_number - 3,
      "speaker": "speaker_name",
      "content": "Copy the utterance provided in the input, three turns before the first generated turn."
    },
    {
      "turn_number": first_generated_turn_number - 2,
      "speaker": "speaker_name",
      "content": "Copy the utterance provided in the input, two turns before the first generated turn."
    },
    {
      "turn_number": first_generated_turn_number - 1,
      "speaker": "speaker_name",
      "content": "Copy the utterance provided in the input, one turn before the first generated turn."
    }
  ],
  "generated_turns": [
    {
      "turn_number": first_generated_turn_number,
      "speaker": "next_speaker",
      "content": "The response generated by the specified speaker in this turn, based on the conversation history and conversation metadata."
    },
    {
      "turn_number": first_generated_turn_number + 1,
      "speaker": "next_speaker",
      "content": "The response generated by the specified speaker in this turn, based on the conversation history and conversation metadata."
    },
    ...
  ]
}
```
\end{lstlisting}
\caption{System prompt of the vanilla prompt for conversation continuation.}
\label{prompt:generation-system}
\end{table*}

\begin{table*}[t]
\begin{lstlisting}[style=codebox]
Your task is to generate 1 responses from turn 295 to turn 295.

## Conversation Metadata

{
    "task_goal": "Design, evaluate, and finalize a prototype for a new remote control device by reviewing its form, features, cost, and market fit, and agree on next steps for production and marketing.",
    "task_category": "project_planning",
    "org_context": "A cross-functional product development meeting at an electronics company bringing together design (UI and industrial), marketing, and project management to review a clay prototype and ensure it meets budget, usability, and branding requirements.",
    "participants": [
        {
            "id": "Industrial Designer", "name": "Baba", "role": "Industrial Designer", "department": "Design", "reporting_to": "Project Manager", "expertise_level": "expert",
            "expertise_areas": ["industrial design", "product prototyping"],
            "hidden_information": {
                "communication_style": {
                    "length": "8-20 words",
                    "style_types": ["informal", "elaborative", "technical"],
                    "personality_traits": ["analytical", "creative"],
                    "details": "Tends to speak in short, direct statements with occasional technical terms ('double curve', 'base section'). Uses fillers like 'yeah' or 'okay' to confirm. Focuses on physical and material aspects, often suggesting improvements to texture, shape, and feasibility."
                },
                "hidden_agenda": "Ensure the prototype is aesthetically pleasing, physically feasible, and aligns with design integrity.",
                "knowledge_before_conversation": "Deep understanding of materials, manufacturing processes, and prototype constraints; familiar with industrial design principles.",
                "emotion_before_conversation": "Neutral to excited about presenting the design.",
                "posture": "Open and collaborative, but detail-focused and pragmatic.",
                "beliefs_about_others": {
                    "Project Manager": "An authoritative facilitator who will enforce budget and schedule constraints and guide the process.",
                    "Marketing": "Trend-driven and cost-sensitive, likely to question design choices based on market appeal.",
                    "User Interface": "Creative colleague focusing on user-facing features but less attuned to material and manufacturing considerations."
                }
            }
        },

[omitted]

}

## Summary of the Conversation

The Project Manager opened a detailed design review of a clay prototype for a new remote control, calling on David Jordan (User Interface Designer) to present and Baba (Industrial Designer) to discuss feasibility. Jordan revealed a "mushroom"-inspired form in company colors (yellow and red), combining a circular "ball" channel selector, fuzzy finish, triangular base, and mushroom-shaped buttons. Participants joked about its fruit versus vegetable resemblance (lemon, pineapple, mushroom) and noted risks of it being mistaken for food.

Baba explained the internals-a low-level chip, LCD button module, base-mounted battery, and an innovative external solar receptor. Jordan added that voice recognition is built into a microphone array. Marketing pressed on price impact, and the team agreed to verify solar component cost later.

Marketing then led a feature evaluation using a 1-7 scale (1 = true, 7 = false). The first criterion, "fancy look and feel," drew comments that the colors and shape are unique but not sufficiently "fancy," resulting in a rating of 3 (acceptable but room for improvement). Design suggestions included refining texture for tactility, reducing red prominence, and emphasizing the fruit motif (e.g., lemon rind or pineapple texture).

The conversation paused as the group prepared to assess additional criteria-technological innovation, usability, feasibility-and to estimate production costs against the 12.50 budget target. Next steps are to finalize feature ratings, review cost implications (especially for the solar receptor), and plan prototype refinements before moving toward production and marketing plans.

## Recent Conversation History

[
    {
        "turn_number": 265,
        "speaker": "Industrial Designer",
        "content": "It's a microphone array, okay."
    },

[omitted]

    {
        "turn_number": 294,
        "speaker": "Project Manager",
        "content": "It's better like that, isn't it?"
    }
]
\end{lstlisting}
\caption{User prompt of the vanilla prompt for conversation continuation.}
\label{prompt:generation-user}
\end{table*}

\begin{table*}[t]
\begin{lstlisting}[style=codebox]
### Additional Instructions for Inconsistent and Uncollaborative Behaviors

Human conversations may involve inconsistent and uncollaborative behaviors, arising either intentionally or unintentionally. By referring to the conversation history and metadata, identify possible inconsistent and uncollaborative behaviors that may arise in the dialogue, and incorporate such behaviors naturally into the simulated conversation when appropriate. The following list outlines representative inconsistent and uncollaborative behaviors:

#### Possible Inconsistent Behaviors

* Logical Contradiction: Statements that directly contradict earlier statements by the same speaker in the conversation history or Generated Turns, or that contradict the speaker's information in the metadata. For example, if a participant first states they are allergic to peanuts and later says they enjoy eating peanut butter.
* Misunderstanding: Instances where a participant misinterprets or misunderstands information provided earlier in the conversation. For example, if one participant says they will arrive at 3 PM and another participant responds as if they said 5 PM.
* Factual Inaccuracy: Statements that are factually incorrect based on widely accepted knowledge. For example, if a participant claims that the capital of France is Berlin.
* Repetition: Instances where a participant unnecessarily repeats information that has already been clearly stated in the conversation. For example, if one participant says "I will bring the documents" and another participant later says "So you will bring the documents, right?" without any new context.
* Redundant Information: Providing overly long responses that do not add value to the conversation.

#### Possible uncollaborative Behaviors

* Under-answering: Responses that do not fully address the questions or try to evade providing complete information.
* Unclear Intent: Responses that are ambiguous or do not clearly convey the participant's intentions. For example, if a participant responds with "Maybe" when asked a direct question about their plans.
* Off-topic Responses: Statements that diverge significantly from the main topic of the conversation without a clear reason. For example, if a discussion about project deadlines suddenly shifts to unrelated personal anecdotes.
* Interruption: Instances where a participant cuts off another participant mid-sentence or speaks over them in a way that disrupts the flow of conversation. For example, if one participant is explaining something and another participant interjects before they finish.
* Persistent disagreement: When a participant continues to disagree with another participant's point of view despite clear evidence or consensus in the conversation.
\end{lstlisting}
\caption{Part of the system prompt for the taxonomy-guided prompt, which is added to the vanilla prompt.}
\label{prompt:taxonomy-guided}
\end{table*}

\begin{table*}[t]
\begin{lstlisting}[style=codebox]
Your task is to evaluate the consistency and collaborativeness of Generated Turns, a simulated multi-party conversation generated by LLMs. We provide a history ("Conversation History") of the conversation, which is provided to a response generation model as a context, and the continued turns generated by the model.

If you do not assign a score of 10, your explanation must clearly specify the reasons why the generated turns did not merit a perfect score with specific examples from the conversation history and the generated turns.

## Evaluation Criteria

**Consistency**

Consistency ("consistency") refers to whether the generated turns are internally consistent and whether they align logically with the provided metadata and the overall conversation history. The evaluation should assess whether the generated turns preserve logical flow and relevance to the preceding conversation. Responses should not be penalized for length or phrasing if they remain logically coherent. Provide a detailed analysis of the logical relationships both among the generated turns themselves and between those turns and the prior conversation context, including any implicit premises, dependencies, or inferences that should be maintained.

* 1-3: The response shows severe logical failure. It contradicts prior turns, metadata, or other generated turns, introduces irrelevant or factually incompatible content, or ignores critical contextual information. Logical relations are missing both within the generated turns and between them and the conversation history or metadata, resulting in incoherence.
* 3-5: The response partially addresses the topic but breaks logical continuity. It may misinterpret earlier turns, skip essential reasoning steps, or respond inappropriately to the dialogue act (for example, answering a statement or ignoring a question). Dependencies across generated turns or with metadata and history are broken, causing inconsistencies or reasoning gaps.
* 5-6: The response maintains basic logical order but lacks depth or precision. It stays on topic and avoids direct contradictions, yet misses one or more implicit premises, simplifies or overlooks causal or temporal logic, or exhibits weak internal coherence among generated turns. The reasoning chain is incomplete or loosely connected to metadata or prior context.
* 7: The response correctly follows the main explicit premises and maintains logical validity but fails to incorporate secondary dependencies such as implied goals, indirect constraints, or assumptions. Minor internal inconsistencies may occur between generated turns or with metadata, resulting in an incomplete logical representation of the conversation context.
* 8: The response handles both explicit and most implicit premises accurately. It preserves internal consistency among generated turns and aligns logically with conversation history and metadata. Minor lapses, such as omission of subtle contextual cues, partial causal links, or unaddressed edge conditions, slightly weaken the overall logical precision.
* 9: The response demonstrates precise logical continuity. It fully integrates both explicit and implicit reasoning steps, resolves all dependencies (e.g., causal, temporal, conditional), and reflects accurate understanding of participant intentions. Only minor redundancy or over-specification may prevent it from scoring a 10.
* 10: The response achieves complete logical alignment. It is internally consistent across all generated turns and fully coherent with the conversation history and metadata. All explicit and implicit premises, dependencies, and reasoning chains are correctly maintained, producing a contextually optimal and logically flawless continuation.

**Collaborativeness**

Collaborativeness ("collaborativeness") refers to how well the generated turns demonstrate cooperative engagement among participants. The evaluation should assess whether participants build on each other's contributions, maintain logical and factual agreement, and work together toward shared objectives without introducing unwarranted disagreement or contradiction.

* 1-3: The response exhibits active conflict or contradiction among participants. Turns directly oppose or undermine each other's statements or goals without justification, introduce irrelevant disagreement, or disrupt cooperative progress. The conversation shows no sense of shared purpose.
* 3-5: The response shows limited awareness of collaboration. Participants occasionally acknowledge each other but frequently introduce inconsistencies, dismiss prior reasoning, or pursue divergent directions. Cooperative alignment is weak or fragmented, resulting in partial or stalled progress.
* 5-6: The response maintains surface-level agreement and avoids overt conflict, but collaboration is minimal. Participants operate largely in parallel, with limited evidence of building upon one another's ideas or reasoning. Cooperative structure exists but lacks meaningful integration.
* 7: The response shows general cooperative behavior and maintains logical alignment among participants. However, it may overlook opportunities for deeper coordination, refinement, or joint reasoning. Minor inconsistencies or weak support across turns slightly reduce overall coherence.
* 8: The response reflects strong collaboration. Participants acknowledge and extend each other's ideas, resolve minor inconsistencies effectively, and maintain coherent progress toward shared goals. Small gaps in coordination or limited integration of reasoning may remain.
* 9: The response demonstrates effective and sustained collaboration. All participants engage constructively, align on shared reasoning, and build cohesively on prior turns. Dependencies and agreements are fully maintained, with only negligible inefficiencies in cooperative structure.
* 10: The response achieves complete collaborative alignment. All participants consistently support and extend one another's reasoning, maintain full logical and factual agreement, and jointly advance toward shared objectives with no contradictions or missed opportunities for cooperation.
\end{lstlisting}
\caption{Part of the system prompt for LLM-as-a-Judge for the overall evaluation.}
\label{prompt:overall-evaluation}
\end{table*}

\begin{table*}[t]
\begin{lstlisting}[style=codebox]
Your task is to evaluate the consistency and cooperativeness of Generated Turns, a simulated multi-party conversation generated by LLMs. We provide a history ("Conversation History") of the conversation, which is provided to a response generation model as a context, and the continued turns generated by the model.
You will count the number of utterances in the Generated Turns that include the specified inconsistent or uncollaborative behavior.

### Inconsistent Behaviors to Look For

* Logical Contradiction: Statements that directly contradict earlier statements by the same speaker in the conversation history or Generated Turns, or that contradict the speaker's information in the metadata. For example, if a participant first states they are allergic to peanuts and later says they enjoy eating peanut butter.
* Misunderstanding: Instances where a participant misinterprets or misunderstands information provided earlier in the conversation. For example, if one participant says they will arrive at 3 PM and another participant responds as if they said 5 PM.
* Factual Inaccuracy: Statements that are factually incorrect based on widely accepted knowledge. For example, if a participant claims that the capital of France is Berlin.
* Repetition: Instances where a participant unnecessarily repeats information that has already been clearly stated in the conversation. For example, if one participant says "I will bring the documents" and another participant later says "So you will bring the documents, right?" without any new context.
* Redundant Information: Providing overly long responses that do not add value to the conversation.

### Uncollaborative Behaviors to Look For

* Under-answering: Responses that do not fully address the questions or try to evade providing complete information.
* Unclear Intent: Responses that are ambiguous or do not clearly convey the participant's intentions. For example, if a participant responds with "Maybe" when asked a direct question about their plans.
* Off-topic Responses: Statements that diverge significantly from the main topic of the conversation without a clear reason. For example, if a discussion about project deadlines suddenly shifts to unrelated personal anecdotes.
* Interruption: Instances where a participant cuts off another participant mid-sentence or speaks over them in a way that disrupts the flow of conversation. For example, if one participant is explaining something and another participant interjects before they finish.
* Persistent disagreement: When a participant continues to disagree with another participant's point of view despite clear evidence or consensus in the conversation.

## Output Format

Output the results in the following JSON format. For each type of inconsistent or uncollaborative behavior, provide a brief explanation on which utterances were counted, along with the total count of such utterances in the Generated Turns. Then, in the following item, provide the number of utterances that exhibited that behavior. Each count should be in a range of 0 to N, where N is the total number of utterances in the Generated Turns.

```json
{
    "logical_contradiction_explanation": "Explanation of the logical contradiction found.",
    "logical_contradiction_count": an integer representing the number of utterances with logical contradictions,
    "misunderstanding_explanation": "Explanation of the misunderstanding found.",
    "misunderstanding_count": an integer representing the number of utterances with misunderstandings,
    "factual_inaccuracy_explanation": "Explanation of the factual inaccuracy found.",
    "factual_inaccuracy_count": an integer representing the number of utterances with factual inaccuracies,
    "repetition_explanation": "Explanation of the repetition found.",
    "repetition_count": an integer representing the number of utterances with repetitions,
    "redundant_information_explanation": "Explanation of the redundant information found.",
    "redundant_information_count": an integer representing the number of utterances with redundant information,
    "under_answering_explanation": "Explanation of the under-answering found.",
    "under_answering_count": an integer representing the number of utterances with under-answering,
    "unclear_intent_explanation": "Explanation of the unclear intent found.",
    "unclear_intent_count": an integer representing the number of utterances with unclear intent,
    "off_topic_explanation": "Explanation of the off-topic responses found.",
    "off_topic_count": an integer representing the number of utterances with off-topic responses,
    "interruptions_explanation": "Explanation of the interruptions found.",
    "interruptions_count": an integer representing the number of utterances with interruptions,
    "persistent_disagreement_explanation": "Explanation of the persistent disagreement found.",
    "persistent_disagreement_count": an integer representing the number of utterances with persistent disagreement
}
```
\end{lstlisting}
\caption{Part of the prompt for LLM-as-a-Judge for fine-grained evaluation.}
\label{prompt:fine-grained-evaluation}
\end{table*}

\end{document}